\documentclass[10pt,twocolumn,letterpaper]{article}

\usepackage[pagenumbers]{iccv} 

\definecolor{iccvblue}{rgb}{0.21,0.49,0.74}
\usepackage[pagebackref,breaklinks,colorlinks,allcolors=iccvblue]{hyperref}

\usepackage{float}
\usepackage{amsmath} 
\usepackage{cuted}
\usepackage{booktabs}
\usepackage{times}
\usepackage{epsfig}
\usepackage{graphicx}
\usepackage{amsmath}
\usepackage{array}
\usepackage{xcolor}
\def\paperID{6838} 
\def\confName{ICCV}
\def\confYear{2025}

\def\paper{Diff$^2$I2P}
\def\csd{Control-Side Score Distillation}
\def\dct{Deformable Correspondence Tuning}
\def\dd{differentiable densification}

\title{
\textcolor{iccvblue}{Diff$^{\mathbf{2}}$}I2P: \textcolor{iccvblue}{Diff}erentiable Image-to-Point Cloud Registration with
\textcolor{iccvblue}{Diff}usion Prior
} 

\author{
Juncheng Mu\textsuperscript{\rm 1,2$^{*}$}~
Chengwei Ren\textsuperscript{\rm 1,2$^{*}$}~
Weixiang Zhang\textsuperscript{\rm 1}~
Liang Pan\textsuperscript{\rm 2$^{\dagger}$}~
Xiao-Ping Zhang\textsuperscript{\rm 1$^{\dagger}$}~
Yue Gao\textsuperscript{\rm 1$^{\dagger}$}
\\
\small{\textsuperscript{\rm 1}Tsinghua University}, \small{\textsuperscript{\rm 2}Shanghai AI Laboratory} \\
{\tt\scriptsize\{mujc21, rcw22, zhang-wx22\}@mails.tsinghua.edu.cn, paul007pl2020@gmail.com, xpzhang@ieee.org, gaoyue@tsinghua.edu.cn}\\
\small{\textcolor{gray}{$^{*}$ Equal contributions \quad $^{\dagger}$ Corresponding authors}}\\
}

\begin{document}
\maketitle
\begin{abstract}

Learning cross-modal correspondences is essential for image-to-point cloud (I2P) registration.
Existing methods achieve this mostly by utilizing metric learning to enforce feature alignment across modalities, disregarding the inherent modality gap between image and point data. Consequently, this paradigm struggles to ensure accurate cross-modal correspondences.
To this end, inspired by the cross-modal generation success of recent large diffusion models, we propose \textbf{\paper}, a fully \textbf{Diff}erentiable \textbf{I2P} registration framework, leveraging a novel and effective \textbf{Diff}usion prior for bridging the modality gap.
Specifically, we propose a \csd\ (CSD) technique to distill knowledge from a depth-conditioned diffusion model to directly optimize the predicted transformation. However, the gradients on the transformation fail to backpropagate onto the cross-modal features due to the non-differentiability of correspondence retrieval and PnP solver. To this end, we further propose a \dct\ (DCT) module to estimate the correspondences in a differentiable way, followed by the transformation estimation using a differentiable PnP solver. With these two designs, the Diffusion model serves as a strong prior to guide the cross-modal feature learning of image and point cloud for forming robust correspondences, which significantly improves the registration.
Extensive experimental results demonstrate that \paper\ consistently outperforms SoTA I2P registration methods, achieving over 7\% improvement in registration recall on the 7-Scenes benchmark.
Code will be available at \href{https://github.com/mujc2021/Diff2I2P}{https://github.com/mujc2021/Diff2I2P}.

\end{abstract}

\section{Introduction}
\label{sec:intro}

Cross-modal registration between images and point clouds is a crucial task in computer vision with broad applications in robotics, AR/VR, etc. Given an image and a point cloud of the same scene, the goal is to estimate a rigid transformation that aligns the point cloud with the image's camera coordinate system. Unlike single-modal registration, such as image registration \cite{detection_free1,detection_free2,detection_free3,detection_free4,detection_free5,detection_free6} and point cloud registration \cite{smooth, 3dmatch, d3feat, predator, cofinet, geotransformer}, which have been extensively studied for decades, cross-modal registration encounters greater challenges due to limited overlap, severe noise, modality misalignment, etc.

\begin{figure}[t]
\centering
\includegraphics[width=\columnwidth]{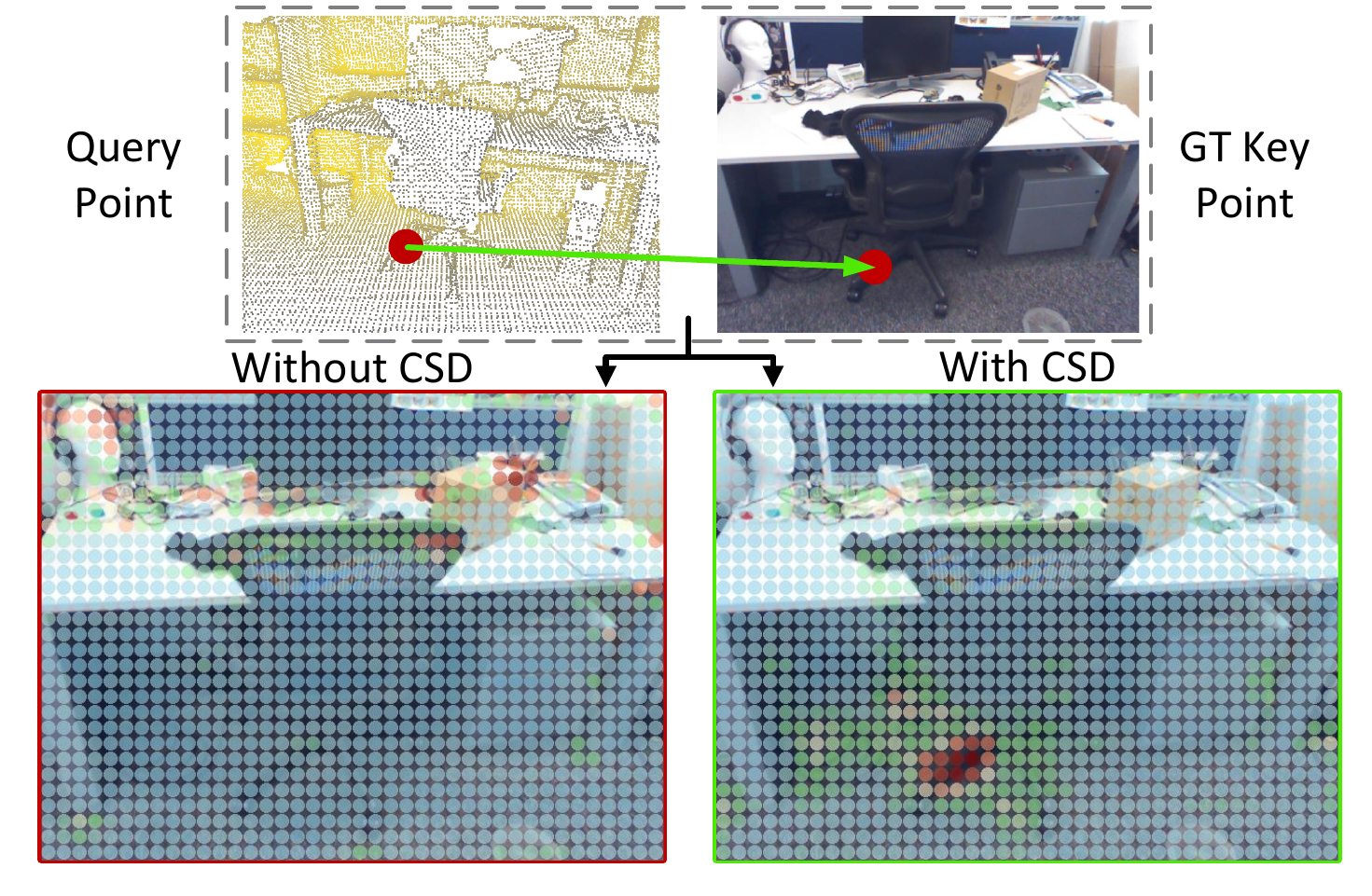}
\vspace{-6mm}
\captionof{figure}{
Our proposed \textbf{\csd\ (CSD)} effectively promotes the learning of cross-modal features by introducing a novel Diffusion prior to bridging the modality gap. To illustrate this, we select a query point from the \emph{chair leg} in the point cloud, where the features are primarily geometry-dominated and challenging for the image backbone to capture effectively. Its ground-truth corresponding pixel is marked in the top right image. The bottom row visualizes the comparison of cosine similarity between image and point cloud features with and without CSD. With CSD, it correctly locates the pixel with the most similar image features while the one without it fails.
}
\vspace{-12pt}
\label{fig:feature}
\end{figure}

Most existing approaches \cite{deepi2p, corri2p, 2d3d-matchnet, lcd, lcd3, p2net, 2d3dmatr} follow the technical roadmap of image or point cloud registration, i.e., \emph{matching and transformation}. They first extract a set of cross-modal correspondences, then the transformation estimation problem is solved as a Perspective-n-Point (PnP) problem using PnP-RANSAC \cite{epnp, ransac}. Therefore, retrieving an accurate putative correspondence set is essential for robust registration. Recent advances \cite{2d3d-matchnet, p2net, 2d3dmatr, differentiable} have led to substantial progress in learning-based matching methods, which typically employ metric learning techniques, such as contrastive and triplet losses, to enforce alignment between image and point cloud features. 

Despite the rapid progress in learning-based cross-modal registration, a notable obstacle remains hindering the performance of current methods, the \textbf{modality gap}. Since feature extraction methods like \cite{resnet,densenet} for 2D images and \cite{pointnet,pointnet++,kpconv} for 3D points mostly only focus on learning local features within their own modality, makes it difficult for the 2D backbones to effectively learn the 3D geometric features from the image. Conversely, the colorless nature of the point cloud makes the 3D backbone fail to adequately capture the texture features of the scene. 
However, existing methods predominantly focus on metric learning to forcibly alleviate modality misalignment, overlooking this gap and resulting in limited performance.

\begin{figure}[t]
\centering
\includegraphics[width=\columnwidth]{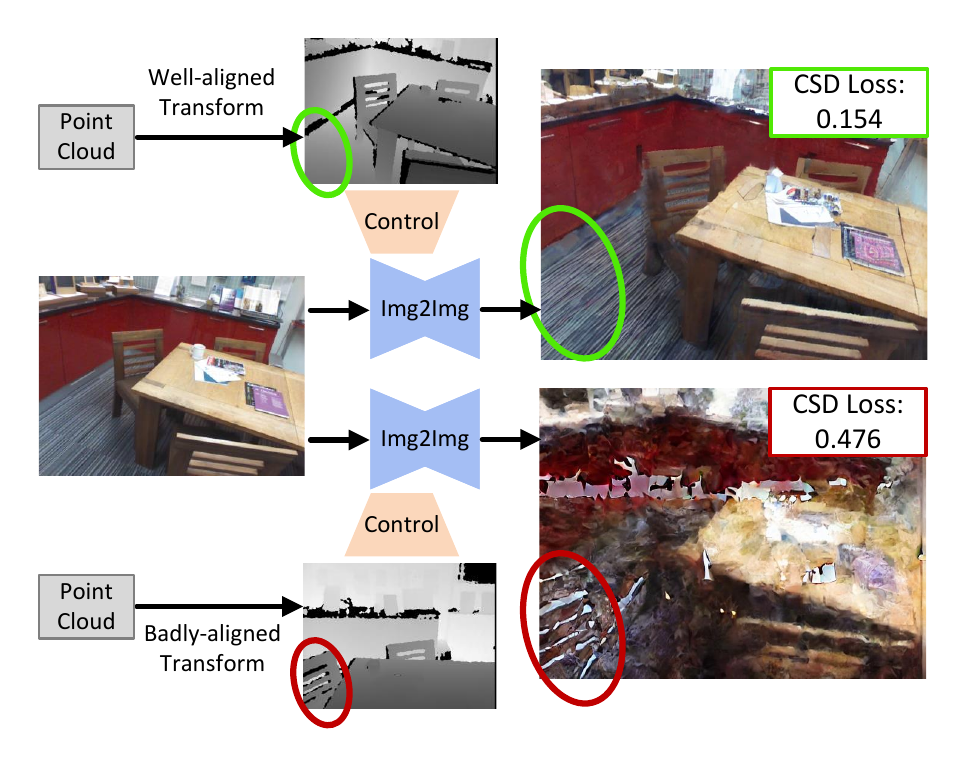}
\vspace{-10mm}
\captionof{figure}{
The illustration of CSD leveraging diffusion prior to effectively evaluate the predicted transformation.
Well-aligned depth and image pairs (top) result in a lower CSD loss and produce clear output images. In contrast, misaligned cases introduce noticeable artifacts (red circle), leading to a higher CSD loss.
}
\label{fig:depth_control}
\vspace{-12pt}
\end{figure}

To tackle these, we propose a novel \csd\ (CSD) technique that distills the 2D texture and 3D geometry knowledge from a depth-conditioned Diffusion \cite{controlnet, ldm} model to promote the cross-modal feature learning for registration. This model takes as input an RGB image with a depth map as the condition, guiding the denoising process by integrating 2D image textures with 3D point geometry to generate a novel high-fidelity image. 
CSD is inspired by the fact that the misaligned depth and image pair will impair the noise prediction capability of this pretrained Diffusion model, leading to severe generation artifacts as shown in \cref{fig:depth_control}. This failure can be modeled by the CSD loss in a Score Distillation Sampling (SDS) manner as follows. The point cloud is first projected into a depth image based on the predicted transformation. Then the input image and depth image are fed into Diffusion for SDS on the ControlNet side. In this way, the alignment can be effectively guided by this cross-modal Diffusion, enabling the image and point cloud backbones to learn distinctive inter-modal features. We illustrate the contribution of CSD in \cref{fig:feature}. The selected query point on the top left contains geometry-dominant features that are difficult for the 2D backbone to learn. The pixel features trained without CSD fail to capture geometry features, leading to ambiguous feature matching. While features trained with CSD effectively model the 3D geometry of the chair leg, achieving accurate and concentrated feature matching. 

Although CSD effectively promotes cross-modal feature learning, the non-differentiability of the correspondence formation process prevents the gradients of transformation from backpropagating to the backbones. Therefore, we further propose a \dct\ (DCT) module, which takes as input the correspondences' features and coordinates and predicts the point offsets for each correspondence. By leveraging cross-modal features, DCT estimates offsets in a differentiable manner while optimizing the alignment of point-pixel pairs. Finally, we utilize a differentiable PnP solver, BPnP \cite{bpnp}, to link the deformed correspondences with CSD by differentiably estimating the transformation. In this way, gradients can flow through the offsets and backpropagate to the backbones. Once the training is complete, we simply discard the Diffusion model and infer sorely using the distilled feature backbones, resulting in fast runtime and low VRAM cost.

Based on these, we propose \paper, a fully differentiable cross-modal registration method capable of bridging the modality gap between image and point cloud. 
Extensive experiments on both 7-Scenes \cite{7-Scenes} and RGB-D Scenes V2 \cite{rgbdscenesv2} benchmarks demonstrate our scene-agnostic superiority. To sum up, our main contributions are three-fold:

\begin{itemize}
    \item We design a fully differentiable image-to-point cloud cross-modal registration method that leverages diffusion prior to bridge the modality gap.
    \item We propose a novel \csd\ (CSD) technique to distill the alignment knowledge from a Depth-conditioned Diffusion for promoting cross-modal feature learning.
    \item We propose a Deformable Correspondence Tuning (DCT) Module to enable differentiable feature matching while refining the correspondence set.
\end{itemize}

\section{Related Work}
\label{sec:related_work}

\begin{figure*}[t]
\centering
\includegraphics[width=\textwidth]{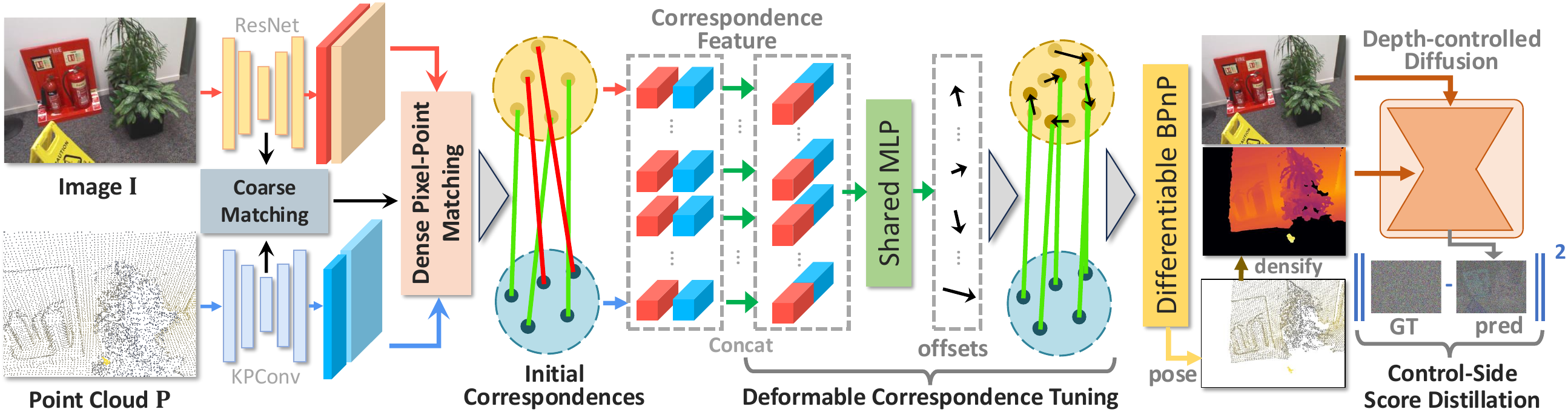}
\captionof{figure}{
\textbf{Pipeline of \paper}. We construct a fully differentiable pipeline for cross-modal registration. Given the input image and point cloud, we first perform feature matching to obtain the initial correspondences. Next, the \dct\ module takes their features as input and predicts point cloud coordinate offsets, ensuring the differentiability meanwhile refining the correspondences. Subsequently, we utilize a BPnP \cite{bpnp} for differentiable transformation estimation and the predicted transform is used to project the point cloud into a depth map. The input image and this depth map are then sent to compute the Control-Side
Score Distillation (CSD) loss for distilling cross-modal features, with its gradient backpropagates through the entire registration network for sufficient optimization.
}
\vspace{-10pt}
\label{fig:pipeline}
\end{figure*}

\textbf{Image-to-Image Registration.} Image registration is of significant importance in computer vision and has been studied for decades. Conventional methods employ keypoint detection approaches by applying handcrafted \cite{ORB, SIFT} or learned \cite{learning-based-descriptor1,learning-based-descriptor2,learning-based-descriptor3,learning-based-descriptor4,learning-based-descriptor5,learning-based-descriptor6,learning-based-descriptor7} descriptors to establish image correspondences. Then, they estimate the transformation using algorithms such as Bundle Adjustment \cite{estimator1,estimator2} or Perspective-n-Point (PnP) \cite{epnp}. Recently, detection-free methods \cite{detection_free1,detection_free2,detection_free3,detection_free4,detection_free5,detection_free6} have also demonstrated promising results.

\noindent
\textbf{Point Cloud Registration.} The development of point cloud registration is similar to image registration. Early methods primarily relied on keypoint detection \cite{ppf,fpfh}, extracting point feature descriptions for registration, such as PPF \cite{ppf} and FPFH \cite{fpfh}. Subsequently, detection-free methods \cite{smooth,3dmatch,d3feat,fcgf,predator} demonstrate greater potential. Recently, some approaches \cite{cofinet,geotransformer,pcr-cg,peal,colorpcr} have begun to employ a coarse-to-fine strategy for registration. They perform feature extraction \cite{pointnet,pointnet++,kpconv} and matching to obtain correspondences, and utilize robust estimators \cite{ransac, dgr, pointdsc} to estimate the transformation.

\noindent
\textbf{Image-to-Point Cloud Registration.} Image-to-point cloud registration \cite{2d3d-matchnet, deepi2p, corri2p, differentiable, diffreg} is more challenging than single-modal registration.
2D3DMATR \cite{2d3dmatr} employs a coarse-to-fine strategy, using transformers to assist in feature extraction, and solve the transformation with PnP \cite{epnp} solver.
However, this force alignment on cross-modal feature learning is insufficient.
FreeReg \cite{freereg} utilizes the Diffusion model to extract cross-modal correspondences, allowing for registration without the need for training. However, it directly applies the Diffusion model for feature extraction, leading to extremely high runtime and computational cost.
Our \paper\ significantly improves the registration accuracy by bridging the modality gap, while maintaining fast inference times.

\noindent
\textbf{Score Distillation Sampling.} Score Distillation Sampling (SDS) \cite{dreamfusion} is widely used in 3D asset generation tasks. Typically, Neural Radiance Fields (NeRF) \cite{nerf} are employed to represent 3D objects, with SDS distilling knowledge from pre-trained diffusion models \cite{ldm} for generation. Some methods \cite{magic3d,fantasia3d,dreamgaussian} further leverage SDS to achieve improved generation results. With the emergence of ControlNet \cite{controlnet}, we draw inspiration from SDS to propose a novel Control-Side Score Distillation technique, which distills knowledge from pre-trained ControlNet to optimize control-side (e.g. depth conditioning control) parameters.
\section{Method}
\label{sec:method}


\subsection{Overview}

Given a pair of 2D image $ \mathbf{I} \in \mathbb{R}^{H \times W \times 3} $ and a 3D point cloud $ \mathbf{P} \in \mathbb{R}^{N \times 3} $, the goal of cross-modal image-to-point cloud registration is to predict a rigid transformation $ \mathcal{T} = [\mathbf{R} | \textbf{t}] $, where $ \mathbf{R} \in \mathcal{SO}^3 $ is the 3D rotation matrix and $ t \in \mathbb{R}^3 $ is the 3D translation vector. The most common pipeline is correspondence-based, which involves finding correspondences between a set of points and pixels, followed by minimizing the 2D projection error between them:
\begin{equation}
    \min _{\mathbf{R}, \mathbf{t}} \sum_{\left(\mathbf{x}_i, \mathbf{y}_i\right) \in \Bar{\mathcal{C}}}\left\|\mathcal{K}\left(\mathbf{R} \mathbf{x}_i+\mathbf{t}; \mathbf{K}\right)-\mathbf{y}_i\right\|^2 ,
\end{equation}
where $ \Bar{\mathcal{C}} = \{ (\mathbf{x}_i, \mathbf{y}_i) | \mathbf{x}_i \in \mathbf{P}, \mathbf{y}_i \in \mathbf{I} \} $ is the correspondences set, $ \mathbf{K} $ is the camera intrinsic matrix, and $ \mathcal{K}: \mathbb{R}^3 \to \mathbb{R}^2 $ is the projection from 3D point space to 2D image plane. This is a PnP problem and can be solved by algorithms like RANSAC \cite{ransac} and EPnP \cite{epnp}. 


However, following the process above, previous work suffers from inaccurate correspondences due to two aspects.
\textbf{First}, the registration process is non-differentiable due to the \textit{argmax} operator used in feature matching and the PnP-RANSAC method in transformation estimation. These prevent the direct supervision of the transformation for sufficient optimization.
\textbf{Second}, these methods typically apply metric learning to forcibly align cross-modal feature spaces.
The inherent modality gap makes it challenging for backbones to effectively learn cross-modal features.
To takle these, we propose a novel module called Deformable Correspondence Tuning (DCT) cooperating with a differentiable PnP solver to enable differentiable registration while refining the impaired correspondences. Then we further propose a Control-Side Score Distillation (CSD) loss leveraging a novel cross-modal Diffusion prior to bridge the modality gap.
The overall pipeline is shown in \cref{fig:pipeline}.

\begin{figure*}[t]
\centering
\includegraphics[width=\textwidth]{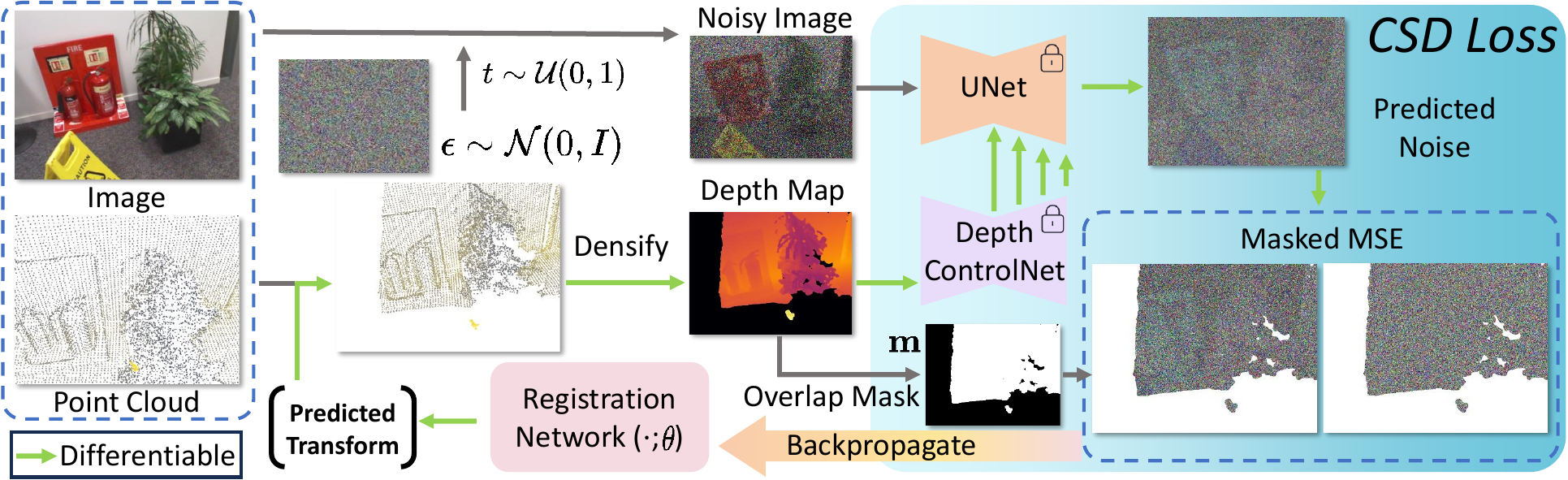}
\vspace{-5mm}
\captionof{figure}{
Illustration of \csd. We employ a depth-conditioned Diffusion model \cite{controlnet, ldm} to bridge the modality gap between images and point clouds. It effectively distills knowledge from the ControlNet to promote cross-modal feature learning for accurate registration. Its gradient flows through the differentiable path marked in green lines to efficiently optimize the registration network.
}
\vspace{-10pt}
\label{fig:csd}
\end{figure*}

\subsection{Differentiable Registration}
\noindent
\textbf{Pixel-Point Matching.} 
We perform multi-scale pixel-point matching in a coarse-to-fine manner following 2D3D-MATR \cite{2d3dmatr}.
Specially, we organize the multi-scale features of the point cloud and image in an FPN \cite{fpn} style, with KPconv \cite{kpconv} and ResNet \cite{resnet} employed to extract features from the point cloud and image. Then we adopt a multi-scale patch matching strategy to alleviate the scale mismatching problem. After selecting the point-pixel pairs with \textit{topk} similarity, we obtain the initial correspondence set $\mathcal{C}_\text{init} = \{ (\mathbf{x}_i, \mathbf{y}_i) | \mathbf{x}_i \in \mathbf{P}, \mathbf{y}_i \in \mathbf{I} \}$.

\noindent
\textbf{Deformable Correspondence Tuning.} 
Inspired by the deformable convolution \cite{dai2017dcn_v1, zhu2019dcn_v2, xiong2024dcn_v4, kpconv} and attention \cite{zhu2020deformable},
our key idea of Deformable Correspondence Tuning (DCT) is to introduce learnable point offsets $ \Delta\mathbf{p}$ to refine the correspondences. These offsets are predicted using the features from both image and point cloud, which direct the gradient of the correspondence to the feature backbones. This not only preserves the gradient flow for differentiable registration but also fine-tunes the correspondences to prevent potential erroneous matches.
For each predicted correspondence $ c_i = (\mathbf{x}_i, \mathbf{y}_i) \in \mathcal{C}_\text{init} $, DCT predicts a point offset $ \Delta\mathbf{p}_i \in \mathbb{R}^3 $ for $ \mathbf{x}_i $ to make it better aligned with $ \mathbf{y}_i $ under the ground-truth transformation $ \Bar{\mathcal{T}} = [\Bar{\mathbf{R}} | \Bar{t}] $. After that, our final correspondence set $\mathcal{C}$ can be updated by adding the offsets to points' coordinate:
\begin{equation}
    \mathcal{C} = \{ (\mathbf{x}_i + \Delta\mathbf{p}_i, \mathbf{y}_i) | (\mathbf{x}_i, \mathbf{y}_i) \in \mathcal{C}_\text{init} \}.
\end{equation}

To optimize the point offset $\Delta\mathbf{p}_i$, we consider the global positions with features from both modalities of each correspondence. Specifically, we concatenate the fine-level features of $(\mathbf{x}_i, \mathbf{y}_i)$ to enable feature sharing between the point and pixel spaces and then provide explicit position information for them, i.e., $ \mathbf{F}^i = [\mathbf{F}_x^i ; \mathbf{F}_y^i ; \mathbf{x}_i ; \mathcal{K}^{-1}(\mathbf{y}_i; \mathbf{K})] \in \mathbb{R}^{2N+6}$, where $ \mathbf{F}_x^i \in \mathbb{R}^N $ and $ \mathbf{F}_y^i \in \mathbb{R}^N $ are their corresponding features.
Then we simply adopt an MLP to predict the offset with $\Delta\mathbf{p}_i = \text{MLP}(\mathbf{F}^i)$.

DCT aims to predict offsets that move points to positions most likely to align with their corresponding pixels. It alleviates the limitation of feature matching, which only considers feature similarity, by incorporating correspondence correlations in coordinate space.
More importantly, DCT enables the backpropagation of gradients from the correspondences to feature extraction backbones, making the differentiable registration feasible.

\noindent
\textbf{Differentiable PnP Solver.} We employ BPnP \cite{bpnp} as the differentiable solver due to its better numerical stability in the context of our work. BPnP efficiently derives accurate gradients of PnP solver based on the Implicit Function Theorem \cite{bpnp}.
By applying BPnP to correspondence set $\mathcal{C}$, we can estimate the transformation $\mathcal{T}$ in a differentiable and effective manner, enabling accurate registration. Please refer to the supplementary materials for details about BPnP.



\subsection{Control-Side Score Distillation}
\label{method:CSD}
The key design of CSD is to present an effective evaluation that explicitly optimizes the predicted transformation. The most intuitive approach is to simply apply the estimated transformation $ \mathcal{T} $ to the point cloud $ \mathbf{P} $ and compute the mean squared error with the image $ \mathbf{I} $:
\begin{equation}
    \mathcal{L}_{\text{trans}} = \sum_{\left(\mathbf{x}_i, \mathbf{y}_i\right) \in \mathcal{C}} \left\| (\mathbf{R}(\mathbf{x}_i)+\mathbf{t}) - \mathcal{K}^{-1}(\mathbf{y}_i; \mathbf{K})\right\|^2,
\end{equation}
where $ \mathcal{C} = \{ (\mathbf{x}_i, \mathbf{y}_i) | \mathbf{x}_i \in \mathbf{P}, \mathbf{y}_i \in \mathbf{I} \} $ is the ground truth correspondences, $ \mathbf{K} $ is the camera intrinsic matrix, and $ \mathcal{K}^{-1} $ is the projection from 2D image plane to 3D point space. 
\emph{However}, our experiments find that this direct optimization of $ \mathcal{L}_{\text{trans}} $ suffers from \textbf{severe training instability} due to the inherent difficulty of cross-modal training. Misalignment caused by poorly estimated transformations can result in an excessively large $ \mathcal{L}_{\text{trans}} $, leading to convergence failure with gradient explosions. To this end, we discard the ground-truth supervision and instead propose a more effective objective namely CSD loss $ \mathcal{L}_{\text{CSD}} $ to distill cross-modal knowledge from the pretrained Diffusion model.
The specific process of $\mathcal{L}_{\text{CSD}}$ calculation is illustrated in \cref{fig:csd}. We first apply the predicted transformation to the point cloud and project it to the image plane to get the depth map, which will serve as the input condition for the Diffusion model. However, this depth map is relatively sparse, differing from the dense depth maps used to train the ControlNet \cite{controlnet}. Therefore, we introduce a \dd\ operator \( \mathcal{F} \) to densify the sparse depth map for conditioning:
\begin{equation}
    \mathbf{D} = \mathcal{F}(\mathcal{K}(\mathbf{R} \cdot \mathbf{P}+\mathbf{t}, \mathbf{K})),
\end{equation}
where $ \mathcal{F} $ represents the morphology operations of dilation and erosion to generate the dense depth map $ \mathbf{D} $.

We then organize the noisy latent $\mathbf{z}_t$ by encoding the image $ \mathbf{I} $ into latent $ \mathbf{z} $ using VAE \cite{vae} and adding a sampled noise $\epsilon$ from $ \mathcal{N}(\mathbf{0}, \mathbf{I}) $ to it with a timestep $t$ controlling the noise strength. Then $\mathbf{z}_t$ along with the dense depth map $ \mathbf{D} $ are fed into the noise predictor  $\hat{\epsilon}_{\phi}$ to predict the added noise. Since we use a depth-conditioned Diffusion model, here $\hat{\epsilon}_{\phi}$ represents the original UNet with ControlNet.
Following SDS \cite{dreamfusion}, we first define a weighted denoising score matching objective \( \mathcal{L}_{\text{Diff}} \) to compute the loss between the predicted noise with the ground truth,
\begin{equation}
\resizebox{0.88\columnwidth}{!}{
    $\mathcal{L}_{\text{Diff}}(z_t, \mathbf{D}) = \mathbb{E}_{t \sim \mathcal{U}(0,1)}^{\epsilon \sim \mathcal{N}(0, I)} \left[ w(t) \left\| \mathbf{m}\circ(\hat{\epsilon}_{\phi} \left( z_t, \mathbf{D};t \right) - \epsilon) \right\|_{2}^{2} \right]$
    },
\label{eqa:l_cdiff}
\end{equation}
where $\phi$ is the learnable parameters of $\hat{\epsilon}_{\phi}$, $w(t)$ is a weighting function that depends on the timestep $t$, and $\circ$ denotes the Hadamard product. $\mathbf{m} \in \{0, 1\}^{h \times w}$ is the overlap mask indicating whether a pixel falls within the non-empty region of the projected depth, where $ h $, $ w $ are the size of latent. 
We exclude the loss from the empty region since there is no effective depth condition to guide the denoising process.

Then its gradient with respect to the parameters of registration network \( \theta \) can be derived as follows:
\begin{align}
    &\nabla_{\theta} \mathcal{L}_{\text{Diff}}(z_t, \mathbf{D}(\theta)) = \\ \nonumber
    &\mathbb{E}_{t, \epsilon} \big[ w(t) \cdot \mathbf{m} \circ \left( \hat{\epsilon}_{\phi} \left( z_t, \mathbf{D} \right) - \epsilon \right) 
    \underbrace{\frac{\mathbf{m} \circ \partial \hat{\epsilon}_{\phi} \left( z_t, \mathbf{D} \right)}{\partial \mathbf{D}}}_{\text{Jacobian}} \frac{\partial \mathbf{D}}{\partial \theta} \big].
\end{align}
Inspired by SDS \cite{dreamfusion}, omitting the above Jacobian $\in \mathbb{R}^{(h\times w)\times (H \times W)}$ helps alleviate the high computational cost due to the massive number of parameters in the UNet and ControlNet \cite{controlnet} of depth-conditioned Diffusion model.
However, the size mismatch between \( \mathbf{D} \in \mathbb{R}^{H \times W} \) and $\epsilon \in \mathbb{R}^{h \times w}$ makes this omission nontrivial in CSD.
To tackle this, we introduce an intermediate variable \( \mathbf{d}(\theta) \in \mathbb{R}^{h \times w} \), which is the resized depth map of $\mathbf{D}$ to match the latent size.
Then $\nabla_{\theta} \mathcal{L}_{\text{Diff}}$ can be reformulated as:
\begin{align}
    &\nabla_{\theta} \mathcal{L}_{\text{Diff}}(z_t, \mathbf{D}(\theta)) = \\ \nonumber
    &
    \resizebox{\columnwidth}{!}{
    $\mathbb{E}_{t, \epsilon} \big[ w(t) \cdot \mathbf{m} \circ \left( \hat{\epsilon}_{\phi} \left( z_t, \mathbf{D} \right) - \epsilon \right) 
    \underbrace{\frac{\mathbf{m} \circ \partial \hat{\epsilon}_{\phi} \left( z_t, \mathbf{D} \right)}{\partial \mathbf{D}} \frac{\partial \mathbf{D}}{\partial \mathbf{d}}}_{\text{size matched Jacobian}} \frac{\partial \mathbf{d}}{\partial \theta} \big].$}
\end{align}
By combining the term $\frac{\partial \mathbf{D}}{\partial \mathbf{d}} \in \mathbb{R}^{(H\times W)\times (h \times w)}$, we get the size matched Jacobian $\in \mathbb{R}^{(h\times w)\times (h \times w)}$, which can be omitted as a unit matrix to save the computation cost.
Our implementation finds that this omission leads to an effective optimization for the registration network \( \theta \) (Sec. \ref{Ablation Studies}).
We formally derive the gradient of $\mathcal{L}_\text{CSD}$ with respect to the parameter of registration network $ \theta $ as:
\begin{align}
\resizebox{0.88\columnwidth}{!}{
    \label{eqa:CSD_loss}
    $\nabla_{\theta} \mathcal{L}_{\text{CSD}}(\mathbf{x}, \mathbf{D}(\theta)) = \mathbb{E}_{t, \epsilon} \left[ w(t) \cdot \mathbf{m} \circ \left( \hat{\epsilon}_{\phi} \left( \mathbf{x}, \mathbf{D};t \right) - \epsilon \right) 
     \frac{\partial \mathbf{d}}{\partial \theta} \right]$}.
\end{align}

During training, we directly compute this \( \nabla_{\theta} \mathcal{L}_{\text{CSD}} \) and backpropagate it to the registration network \( \theta \) for efficient optimization. Please refer to the supplementary materials for more details.

\subsection{Loss functions}

Our model is trained using the sum of three losses, a metric learning loss $ \mathcal{L}_{m} $ for supervising the descriptive image and point cloud features, an offset loss $ \mathcal{L}_{o} $ for supervising the deformable correspondences, and a CSD loss $ \mathcal{L}_\text{CSD} $ for bridging the modality gap, i.e., $ \mathcal{L} = \alpha\mathcal{L}_{m} + \beta\mathcal{L}_{o} + \gamma\mathcal{L}_\text{CSD} $, where $\alpha, \beta$ and $\gamma$ are coefficients of losses.

Following 2D3D-MATR \cite{2d3dmatr}, we utilize $ \mathcal{L}_m $ for correspondences supervision, which is composed of a scaled circle loss $ \mathcal{L}_c $ on coarse-level features and a standard circle loss $ \mathcal{L}_f $ on fine-level features.
In practical implementation, we find that simply extending the original $ \mathcal{L}_c $ from the Transformer to the image and point cloud backbones will augment the learning of distinctive features, so we adopt it for coarse-level supervision. The metric leanrning loss is defined as $ \mathcal{L}_m = \mathcal{L}_\text{aug} + \lambda\mathcal{L}_f $. For detailed information about these losses, please refer to supplementary materials.


The offset loss $ \mathcal{L}_o $ is designed to estimate accurate offsets $ \Delta\mathbf{p}$ for better coordinate tuning. For each $(\mathbf{x}_i, \mathbf{y}_i) \in \mathcal{C}$, we design their distance loss $ \mathcal{L}_{\mathbf{d}}^i $ as:
\begin{equation}
    \label{distance_loss}
    \mathcal{L}_{\mathbf{d}}^i = \left\| (\Bar{\mathbf{R}}(\mathbf{x}_i+\Delta\mathbf{P}_i)+\Bar{\mathbf{t}}) - \mathcal{K}^{-1}(\mathbf{y}_i; \mathbf{K})\right\|^2 ,
\end{equation}
where $ \mathcal{K}^{-1}: \mathbb{R}^2 \to \mathbb{R}^3 $ is the projection function from 2D image plane to 3D point space. To prevent the offsets from compromising feature extraction backbones' capability, we impose an L2-norm regularization to enhance stability, i.e., $\mathcal{L}_\mathbf{r}^i = \left\| \Delta\mathbf{p}_i \right\|_2$. Then, the total offset loss $ \mathcal{L}_o $ can be calculated as,
\begin{equation}
    \mathcal{L}_o = \frac{1}{|\mathcal{C}|} \sum_{i=1}^{|\mathcal{C}|} (\mathcal{L}_\mathbf{d}^i + \mu\mathcal{L}_\mathbf{r}^i ) ,
\end{equation}
where $\mu$ is the coefficient to control regularization strength.

For $ \mathcal{L}_\text{CSD} $, we do not explicitly compute its value. Instead, we directly derive its gradient from \cref{eqa:CSD_loss} for backpropagation.


\section{Experiments}
\label{sec:experiments}
In this section, we first introduce the implementation details (\cref{implementation details}), then we present extensive experimental results of \paper\ and other \textbf{training-based baselines} on 7-Scenes \cite{7-Scenes} (\cref{Evaluations on 7-Scenes}) and RGB-D Scenes V2 \cite{rgbdscenesv2} (\cref{Evaluations on RGB-D Scenes V2}) for evaluation.
In \cref{Ablation Studies}, we conduct extensive ablation studies to validate the contribution of each component.
In \cref{sec:freereg}, we discuss the comparison with another competitive Diffusion-based method FreeReg \cite{freereg}, and present more evaluations on the inference runtime and VRAM.

\begin{figure*}[t]
\centering
\includegraphics[width=\textwidth]{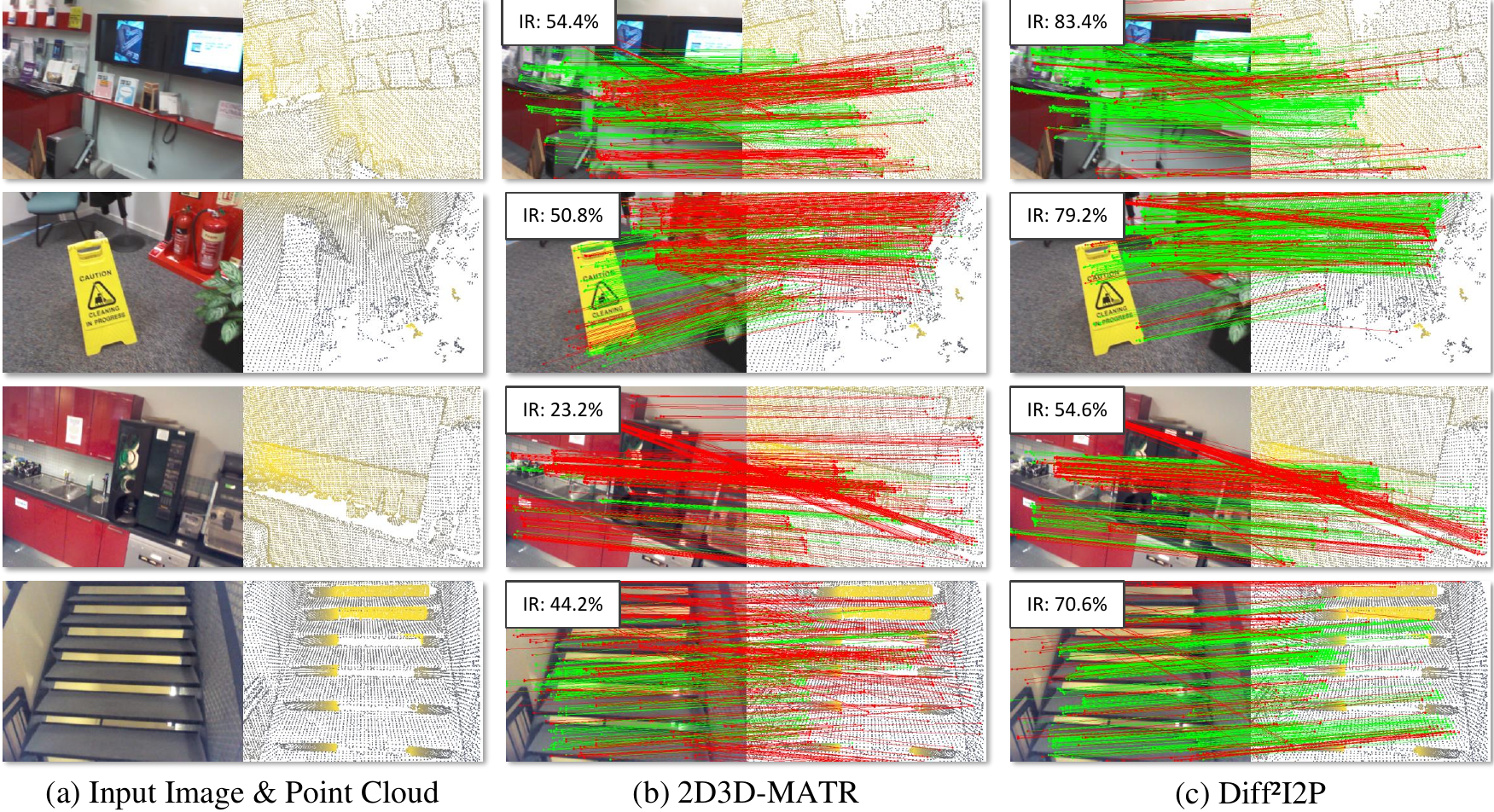}
\setlength{\abovecaptionskip}{-4pt}
\captionof{figure}{
Qualitative results on the 7-Scenes dataset. The red lines indicate erroneous correspondences (3D distance greater than 5 cm), while the green lines represent correct correspondences.
}
\vspace{-8pt}
\label{fig:qualitative}
\end{figure*}

\subsection{Implementation Details}
\label{implementation details}

\noindent
\textbf{Network architecture.} We employ an FPN \cite{fpn} strategy for downsampling and feature extraction of both the image and point cloud, with each modality undergoing four stages of downsampling. For feature extraction, we use ResNet \cite{resnet} for image features and KPConv \cite{kpconv} for point cloud features. The image downsampling reduces the resolution from 480×640 to 60×80, while the point cloud downsampling begins with a voxel size of 2.5 cm, doubling at each following stage. The configuration of the transformer layers is consistent with 2D3D-MATR \cite{2d3dmatr}. The training can be completed with four RTX 3090s within one day.

\noindent
\textbf{Datasets.} We mainly evaluate \paper\ on 7-Scenes \cite{7-Scenes} and RGB-D Scenes V2 \cite{rgbdscenesv2} benchmarks. 7-Scenes \cite{7-Scenes} consists of 7 indoor scenes with 46 RGB-D sequences. Each image-point cloud pair has an overlap of at least 50\%. We utilized the official data split, resulting in 4,048 samples for training, 1,011 for validation, and 2,304 for testing. Our benchmarking on this dataset is consistent with that of \cite{2d3dmatr}. RGB-D Scenes V2 \cite{rgbdscenesv2} includes 14 indoor scenes, comprising 11,427 frames. Each image-point cloud pair has at least 30\% overlap. We randomly split the dataset into training, validation, and test sets, containing 1978, 117, and 386 samples, respectively. To further evaluate \paper\ in dynamic outdoor scenarios, we also provide experiments on KITTI \cite{kitti} in the supplementary material.

\noindent
\textbf{Baselines.} Since our method requires training, we primarily compare \paper\ with four training-based methods, FCGF-2D3D \cite{fcgf}, Predator-2D3D \cite{predator}, P2-Net \cite{p2net}, and 2D3D-MATR \cite{2d3dmatr}. To ensure fair comparisons, the feature extraction backbones used in all comparison methods are kept consistent. 
We also present more discussions and comparisons with another Diffusion-based baseline FreeReg \cite{freereg} in \cref{sec:freereg}.

\noindent
\textbf{Metrics.} We primarily adopt four metrics. Inlier Ratio (IR), which calculates the proportion of correctly estimated pixel-point correspondences (defined as those within 5cm in 3D distance).
Patch Inlier Ratio (PIR), which measures the proportion of correctly estimated patch correspondences (defined as those with overlap bigger than 30\%).
Feature Matching Recall (FMR), which measures the percentage of all point cloud-image pairs with an inlier ratio exceeding a threshold (\textit{i.e.}, 10\%). Registration Recall (RR), which assesses the proportion of all point cloud-image pairs with an RMSE less than a specified threshold (\textit{i.e.}, 10 cm).

\subsection{Evaluations on 7-Scenes}
\label{Evaluations on 7-Scenes}

\begin{table}[t]
\setlength{\abovecaptionskip}{2pt}
\centering
\caption{Registration results of \paper\ and baselines on 7-Scenes. The best of each indicator is in \textbf{bold}.}
\label{tables:7-Scenes}
\resizebox{\columnwidth}{!}{
\large
\begin{tabular}{@{\hspace{0.05cm}}p{1.5cm}@{\hspace{0.9cm}}l@{\hspace{0.08cm}}c@{\hspace{0.11cm}}c@{\hspace{0.13cm}}c@{\hspace{0.18cm}}c@{\hspace{0.27cm}}c@{\hspace{0.15cm}}c@{\hspace{0.08cm}}c@{\hspace{0.05cm}}}

\toprule
Model & Chess & Fire & Heads & Office & Pupk & Kitc & Stairs & Mean \\
\midrule
\multicolumn{9}{c}{Inlier Ratio $\uparrow$} \\
\midrule
FCGF~\cite{fcgf} & 34.2 & 32.8 & 14.8 & 26.0 & 23.3 & 22.5 & 6.0 & 22.8 \\
P2Net~\cite{p2net} & 55.2 & 46.7 & 13.0 & 36.2 & 32.0 & 32.8 & 5.8 & 31.7 \\
Predator~\cite{predator} & 34.7 & 33.8 & 16.6 & 25.9 & 23.1 & 22.2 & 7.5 & 23.4 \\
MATR~\cite{2d3dmatr} & 72.1 & 66.0 & 31.3 & 60.7 & 50.2 & 52.5 & \textbf{18.1} & 50.1 \\
\textbf{Ours}&\textbf{74.1}&\textbf{68.8}&\textbf{39.2}&\textbf{65.6}&\textbf{52.1}&\textbf{54.2}&\textbf{18.1}&\textbf{53.2}\\
\midrule
\multicolumn{9}{c}{Feature Matching Recall $\uparrow$} \\
\midrule
FCGF~\cite{fcgf} & 99.7 & 98.2 & 69.9 & 97.1 & 83.0 & 87.7 & 16.2 & 78.8 \\
P2Net~\cite{p2net} & \textbf{100.0} & 99.3 & 58.9 & 99.1 & 87.2 & 92.2 & 16.2 & 79.0 \\
Predator~\cite{predator} & 91.3 & 95.1 & 76.7 & 88.6 & 79.2 & 80.6 & 31.1 & 77.5 \\
MATR~\cite{2d3dmatr} & \textbf{100.0} & 99.6 & 98.6 & \textbf{100.0} & 92.4 & 95.9 & \textbf{58.1} & 92.1 \\
\textbf{Ours}&\textbf{100.0}&\textbf{100.0}&\textbf{100.0}&\textbf{100.0}&\textbf{93.4}&\textbf{96.2}&55.4&\textbf{92.2}\\
\midrule
\multicolumn{9}{c}{\textbf{Registration Recall} $\uparrow$} \\
\midrule
FCGF~\cite{fcgf} & 89.5 & 79.7 & 19.2 & 85.9 & 69.4 & 79.0 & 6.8 & 61.4 \\
P2Net~\cite{p2net} & 96.9 & 86.5 & 20.5 & 91.7 & 75.3 & 85.2 & 4.1 & 65.7 \\
Predator~\cite{predator} & 69.6 & 60.7 & 17.8 & 62.9 & 56.2 & 62.6 & 9.5 & 48.5 \\
MATR~\cite{2d3dmatr} & 96.9 & 90.7 & 52.1 & 95.5 & 80.9 & 86.1 & 28.4 & 75.8 \\
\textbf{Ours}&\textbf{99.0}&\textbf{95.6}&\textbf{74.0}&\textbf{98.9}&\textbf{86.8}&\textbf{90.2}&\textbf{36.5}&\textbf{83.0}\\
\bottomrule
\end{tabular}
}
\vspace{-8pt}
\end{table}

\noindent
\textbf{Quantitative results.} We train \paper\ on the 7-Scenes \cite{7-Scenes} dataset and the experimental results are shown in \cref{tables:7-Scenes}. For IR, \paper\ consistently outperforms previous methods, indicating more accurate feature matching due to the differentiable supervision with CSD loss. Our method also achieves the best FMR across most scenes, reaching 100\% in four scenes, and outperform the baselines in the most of cases. Regarding the most crucial metric RR, \paper\ shows comprehensive performance improvements, with an average RR exceeding that of 2D3D-MATR \cite{2d3dmatr} by 7.2\% and achieving a 22\% increase in the challenging \textit{Heads} scene. These results demonstrate the breakthrough advancements of our proposed fully differentiable pipeline with CSD supervision.

\noindent
\textbf{Qualitative results.} \cref{fig:qualitative} visualizes the correspondences estimated by 2D3D-MATR and \paper\ on 4 scenes from the 7-Scenes dataset. Column (a) shows the input image and point cloud pairs, while columns (b) and (c) show the correspondences from 2D3D-MATR and our \paper, respectively. We selected the top 500 matching pairs based on their matching scores for visualization. 
The first and second rows depict scenes from \textit{redkitchen} and \textit{fire}, which contain relatively rich colors and clear geometric features. Both methods achieve a high IR in these cases, with \paper\ better identifying regions with prominent textures and geometric features, achieving an IR of approximately 80\%. 
In the third and fourth rows, the scene contains complicated and repetitive geometric patterns under complex lighting, making it challenging for precise feature matching. 2D3D-MATR generates numerous incorrect matches, whereas our method achieves significantly more accurate matching, with an inlier ratio (IR) twice as high in comparison. This indicates the differentiable supervision with CSD loss effectively facilitates the learning of cross-modal features.

\subsection{Evaluations on RGB-D Scenes V2}
\label{Evaluations on RGB-D Scenes V2}

\begin{table}[t]
\setlength{\abovecaptionskip}{2pt}
\centering
\vspace{-0.27cm}
\caption{Registration results of \paper\ and baselines on RGB-D Scenes V2. The best of each indicator is in \textbf{bold}.}
\label{tables:rgbd_scenes_v2}
\footnotesize  
\setlength{\extrarowheight}{-15pt}  
\resizebox{\columnwidth}{!}{
\begin{tabular}{l|cccc}
\toprule
\textbf{Method} & PIR (\%)$\uparrow$ & IR (\%)$\uparrow$ & FMR (\%)$\uparrow$ & \textbf{RR} (\%)$\uparrow$ \\
\midrule
FCGF-2D3D \cite{fcgf} & 20.1 & 10.3 & 29.2 & 32.5 \\
P2-Net \cite{p2net} & 30.4 & 14.5 & 63.7 & 41.7 \\
Predator-2D3D \cite{predator} & 32.6 & 15.8 & 68.1 & 33.6 \\
2D3D-MATR \cite{2d3dmatr} &57.6& 36.3 &76.0&56.9\\
\paper\ (\textbf{ours}) &\textbf{60.8}& \textbf{36.9} & \textbf{77.1} & \textbf{60.5} \\
\bottomrule
\end{tabular}
}
\vspace{-5pt}
\end{table}
We train \paper\ on the RGBD Scenes V2 \cite{rgbdscenesv2} dataset and compare it with the baselines. The experimental results are shown in \cref{tables:rgbd_scenes_v2}.
For intermediate metrics such as PIR, IR, and FMR, \paper\ demonstrates a significant improvement over the baselines, highlighting the effectiveness of feature matching. This improvement benefits from the cross-modal supervision provided by the CSD loss, enabling \paper\ to more accurately estimate the transformation, and consequently achieve a higher registration recall.

\begin{figure}[t]
\centering
\includegraphics[width=\columnwidth]{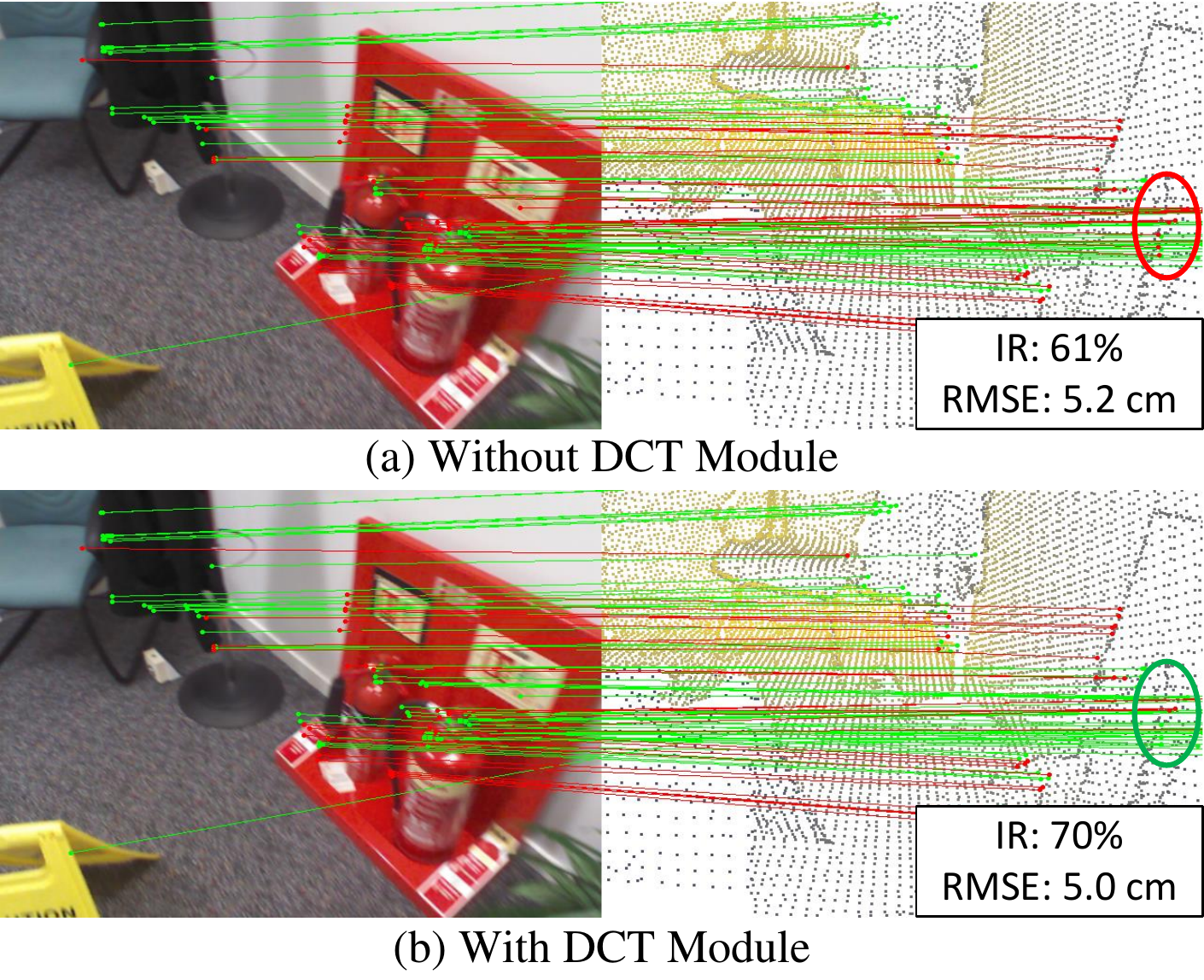}
\vspace{-6mm}
\captionof{figure}{
Illustration of the DCT for refining correspondences. (a) shows the correspondences before offsets refinement. (b) shows the correspondences after offset adjustments. Although DCT can fine-tune the correspondence to improve registration accuracy, we still emphasize that its \emph{principal function} is to ensure \textbf{training differentiability}, i.e., enabling the gradient of the predicted transformation to be backpropagated to the backbone. Without DCT, the distillation of the Diffusion model can not be achieved.
}
\vspace{-8pt}
\label{fig:ablation_dct}
\end{figure}

\subsection{Ablation Studies}
\label{Ablation Studies}
We conduct ablation studies on the network structure of \paper\ to validate the contribution of each component. All experiments are performed on the 7-Scenes \cite{7-Scenes}. 

\noindent
\textbf{\dct.} The absence of the DCT module would disrupt the differentiability of training, we can't ablate it in the training phase.
Therefore, we ablate it during inference instead. The visualization of the estimated correspondences with and without DCT refinement are illustrated in \cref{fig:ablation_dct}. The predicted offsets refine the incorrect correspondences and further improve their overall alignment. Here we emphasize that the \emph{principal function} of DCT is to ensure \textbf{training differentiability}, i.e., enabling the gradient of the predicted transformation to be backpropagated to the backbone. Without DCT, the distillation of the Diffusion model can not be achieved.

\noindent
\textbf{Loss Functions.} We ablate the proposed loss functions, including the circle loss used in \cite{2d3dmatr}, our augmented circle loss, CSD loss, CSD loss without the overlap mask, and the transformation loss. The experimental results are shown in \cref{tables:ablation}, where these losses are abbreviated as \emph{circle}, \emph{aug}, \emph{CSD}, \emph{CSD*}, and \emph{trans}, respectively. Here we retain the offset loss in all experiments for training the DCT module.

In \cref{tables:ablation}, experiment (e) shows the loss combination selected for \paper, which achieves the highest registration recall of 83.0\% on the 7-Scenes dataset. Experiment (a) ablates our augmented circle loss, and compared to (e), all metrics show a decline. This demonstrates the crucial role of the aug-circle loss in providing more comprehensive guidance for correspondence estimation. Experiment (b) uses the transformation loss, which applies the predicted transformation to the point cloud and computes the MSE loss with their corresponding image points. However, this loss is highly unstable, particularly when the transformation bias is large, which can result in an extremely high loss and lead to gradient explosions. Experiments (c) and (d) show the results of canceling the CSD loss and changing to the unmasked version of CSD loss, respectively. Both demonstrate a noticeable degradation in registration recall compared to experiment (e), with other metrics fluctuating slightly. These experiments solidly validate our strategy of loss selection. This further validates that with well-designed architecture and loss design, the strong cross-modal features of the Diffusion model are effectively distilled into our feature backbone, enabling robust and accurate registration.

\begin{table}[t]
\setlength{\abovecaptionskip}{2pt}
\centering
\caption{Ablation study of the loss functions on 7Scenes. CSD$^*$ indicates the CSD without the overlap mask.}
\label{tables:ablation}
\resizebox{\columnwidth}{!}{
\begin{tabular}{c|c@{\hspace{0.1cm}}c@{\hspace{0.1cm}}c@{\hspace{0.1cm}}c@{\hspace{0.1cm}}c|cccc}
\toprule
Type&circle&aug&CSD$^*$&CSD&trans& PIR & IR & FMR & \textbf{RR} \\
\midrule
(a)&    \checkmark&&&\checkmark&    &86.5&53.1&92.0&82.5\\
(b)&    &\checkmark &&& \checkmark& \multicolumn{4}{c}{Gradient Explosion} \\
(c)&    &\checkmark&&&    &86.3&53.0&91.6&81.3\\
(d)&    &\checkmark&\checkmark&&    &86.4&52.6&91.4&81.7\\
(e)&    &\checkmark&&\checkmark&    &\textbf{86.7}&\textbf{53.2}&\textbf{92.1}&\textbf{83.0}\\

\bottomrule
\end{tabular}
}
\vspace{2pt}
\end{table}

\subsection{Comparison with FreeReg}
\label{sec:freereg}
FreeReg \cite{freereg} is another strong baseline based on Diffusion model. It extracts cross-modal correspondence in a training-free manner by utilizing intermediate Diffusion features for matching. The detailed comparison of performance and computational costs is outlined in \cref{tab:freereg_compare}.

\noindent
\textbf{Discussion.}
Despite the inherent unfairness in comparing our method with FreeReg due to the disparity in training requirements, we include it as an informative baseline to highlight the strengths of our method in two key aspects.
\emph{First}, unlike directly using intermediate features for matching, we distill the cross-modal features from the Diffusion model into the backbones. This reduces the randomness during inference, leading to better performance. \emph{Second}, FreeReg requires a full invocation of the large Diffusion model for each inference, resulting in slow inference speed and high VRAM consumption.
In contrast, our method only distill knowledge during training, enabling efficient inference.

\begin{table}[t]
\setlength{\abovecaptionskip}{2pt}
\centering
\vspace{-6pt}
\caption{Comparisons with FreeReg \cite{freereg}.}
\label{tab:freereg_compare}
\footnotesize  
\setlength{\extrarowheight}{-15pt}  
\resizebox{\columnwidth}{!}{
\begin{tabular}{l|cccc}
\toprule
\textbf{Method}& \textbf{RR (\%)$\uparrow$}& \textbf{Time (s)$\downarrow$} & \textbf{Size (MB)$\downarrow$}& \textbf{VRAM (GB)$\downarrow$}\\
\midrule
2D3D-MATR \cite{2d3dmatr} & 75.8 & 0.072 & 118.62 & 3.1\\
FreeReg \cite{freereg} & 71.2 & 9.634 & - & 14.2\\
\paper\ (ours)& \textbf{83.0} & 0.074 & 118.75 & 3.1\\
\bottomrule

\end{tabular}
}
\vspace{-6pt}
\end{table}
\section{Conclusion}
\label{sec:conclusion}
In this paper, we present \paper, a fully differentiable pipeline for cross-modal registration that bridges the modality gap between image and point cloud. With our proposed CSD loss, the training process can distill 2D texture and 3D geometry knowledge from a pretrained depth-controlled Diffusion model for cross-modal feature learning. We further propose a DCT module to introduce differentiability to the correspondence set while refining impaired matches. By combining a differentiable BPnP solver, we construct a fully differentiable network for sufficient optimization. Extensive experiments on both 7-Scenes and RGB-D Scenes V2 benchmarks demonstrate the state-of-the-art performance of \paper, significantly surpassing the strongest baselines. We hope our proposed method inspire further work in differentiable registration and bridging the modality gap. 

\noindent
\textbf{Acknowledgments}. This work was supported by the NSFC (624B1007).

\newpage


{
    \small
    \bibliographystyle{ieeenat_fullname}
    \bibliography{main}
}

\clearpage
\setcounter{page}{1}
\maketitlesupplementary


\section*{Appendix}
\label{sec:appendix}
In this supplementary material, we first provide precise definitions of the evaluation metrics used in the paper (Sec. \hyperref[evaluation_metrics]{A}).
Next, we provide a detailed introduction and discussion of the related work (Sec. \hyperref[baseline discussion]{B}).
Then, we offer a detailed description of the training and test datasets (Sec. \hyperref[datasets]{C}), including their partitioning method. Additionally, we describe the network architecture and implementation details (Sec. \hyperref[implementation_details]{D}). We also conduct additional experiments (Sec. \hyperref[additional_experiments]{E}) such as further metric measurements and runtime analysis. Finally, we present more visualization results on multiple datasets to illustrate the performance of the proposed method intuitively (Sec. \hyperref[visualization]{F}).
\section*{A. Evaluation Metrics}
\label{evaluation_metrics}

\textbf{Inlier Ratio (IR)}: We follow \cite{p2net} to compute the indicator inlier ratio. The Inlier Ratio for cross-modal registration measures the proportion of point-to-pixel correspondences $(\mathbf{x_i}, \mathbf{y_i}) \in \mathcal{C}$ that are within a certain residual threshold under the ground truth transformation $\Bar{\mathcal{T}}_{\mathbf{I}}^{\mathbf{P}}$. Here, $\mathcal{C}$ denotes the estimated correspondence set between the 3D point set $\mathbf{I}$ and the image pixel set $\mathbf{P}$, and $\Bar{\mathcal{T}}_{\mathbf{I}}^{\mathbf{P}}$ represents the ground truth transformation from $\mathbf{I}$ to $\mathbf{P}$. A correspondence pair is considered an inlier if the Euclidean norm of its residual is less than the threshold $\tau_1=10cm$. The Inlier Ratio for the cross-modal pair $\mathbf{I}$ and $\mathbf{P}$ is computed as:

\begin{equation}
\text{IR}(\mathbf{I}, \mathbf{P}) = \frac{1}{|\mathcal{C}|} \sum_{(\mathbf{x}_i, \mathbf{y}_i) \in \mathcal{C}} \mathbb{I} \left[ \|\Bar{\mathcal{T}}_{\mathbf{I}}^{\mathbf{P}}(\mathcal{K}^{-1}(\mathbf{y}_i)) - \mathbf{x}_i\| < \tau_1 \right],
\end{equation}

where $\mathbb{I}[\cdot]$ is the indicator function that counts the number of correspondences with residuals less than the threshold $\tau_1$ and $\mathcal{K}^{-1}$ is a function that unprojects a pixel to a 3D point.

\noindent
\textbf{Feature Matching Recall (FMR)}: The Feature Matching Recall is used to evaluate the result of feature matching by determining the fraction of cross-modal pairs where the Inlier Ratio exceeds a given threshold, $\tau_2 = 5\%$. This metric reflects the probability of accurately recovering the correct transformation using the estimated correspondence set $\mathcal{C}$, typically with the aid of a robust pose estimation algorithm such as RANSAC \citep{ransac}. For a dataset $\mathcal{D}$ containing $|\mathcal{D}|$ cross-modal pairs, the Feature Matching Recall is defined as follows:

\begin{equation}
   \text{FMR}(\mathcal{D}) = \frac{1}{|\mathcal{D}|} \sum_{(\mathbf{I}, \mathbf{P}) \in \mathcal{D}} \mathbb{I}[IR(\mathbf{I}, \mathbf{P}) > \tau_2],
\end{equation}

where $\mathbb{I}[\cdot]$ is the indicator function that counts the number of cross-modal pairs for which the Inlier Ratio exceeds the threshold $\tau_2$. This metric provides insight into the overall robustness and accuracy of the feature matching process across the entire dataset.

\noindent
\textbf{Patch Inlier Ratio (PIR)}: PIR \cite{2d3dmatr} represents the fraction of patch correspondences whose overlap ratios, under the ground-truth transformation $\Bar{\mathcal{T}}_{\mathbf{I}}^{\mathbf{P}}$, are above 0.3. This metric reflects the quality of the estimated patch correspondences. 

The overlap ratio between an image patch $\tilde{\mathbf{y}}_i$ and a point cloud patch (superpoint) $\tilde{\mathbf{x}}_i$ can be calculated in each modality. The definition of point-cloud-side overlap is:
\begin{equation}
    \text{Overlap}_{\mathbf{P}}(\tilde{\mathbf{x}}_i, \tilde{\mathbf{y}}_i) = \frac{1}{|\mathcal{X}_i|}\sum_{\mathbf{x}_i \in \mathcal{X}_i} \mathbb{I} [ \min_{\mathbf{y}_i \in \mathcal{K}^{-1}(\mathcal{Y})}\|\mathbf{x}_i- \mathbf{y}_i \|_2 < \tau_3],
\end{equation}
where $\mathcal{X}_i$ is the up-sampled points of the superpoint $\tilde{\mathbf{x}}_i$, $\tau_3=3.75$cm is the 3D distance threshold, and the definition of image-side overlap is:
\begin{equation}
    \text{Overlap}_{\mathbf{I}}(\tilde{\mathbf{x}}_i, \tilde{\mathbf{y}}_i) = \frac{1}{|\mathcal{Y}_i|}\sum_{\mathbf{y}_i \in \mathcal{Y}} \mathbb{I} [ \min_{\mathbf{x}_i \in \mathcal{K}(\mathcal{X})}\|\mathbf{x}_i- \mathbf{y}_i \|_2 < \tau_4],
\end{equation}
where $\mathcal{Y}_i$ is the up-sampled pixels of the image patch $\tilde{\mathbf{y}}_i$, $\tau_4=8$ pixels is the 2D distance threshold. We take the smaller one of two overlaps and compute the PIR indicator by:
\begin{equation}
\resizebox{\columnwidth}{!}{
   $\text{PIR} = \frac{1}{|\tilde{\mathcal{C}}|} \sum_{(\tilde{\mathbf{x}}_i, \tilde{\mathbf{y}}_i) \in \tilde{\mathcal{C}}} \mathbb{I} \left[ \min\left(\text{Overlap}_{\mathbf{I}}(\hat{\mathbf{x}}_i), \text{Overlap}_{\mathbf{P}}(\hat{\mathbf{y}}_j)\right) > \tau_5 \right],$
   }
\end{equation}
where $\tilde{\mathcal{C}}$ denotes the estimated set of patch correspondences, $\mathbb{I}[\cdot]$ is the indicator function that returns 1 if the condition inside is true and 0 otherwise, and $\tau_5=0.3$ is the overlap threshold.

\noindent
\textbf{Registration Recall (RR)}: The Registration Recall is a metric used to evaluate the accuracy of cross-modal registration between a 3D point cloud and an image. It measures the fraction of image-point cloud pairs for which the Root Mean Square Error (RMSE) is below a certain threshold, denoted as $\tau_6 = 0.1$m. For a dataset $\mathcal{D}$ containing $|\mathcal{D}|$ pairs of image-point cloud pairs, the Registration Recall is defined as follows:

\begin{equation}
   \text{RR}(\mathcal{D}) = \frac{1}{|\mathcal{D}|} \sum_{(\mathbf{I}, \mathbf{P}) \in \mathcal{D}} \mathbb{I}[RMSE(\mathbf{I}, \mathbf{P}) < \tau_6],
\end{equation}

where $\mathbb{I}[\cdot]$ is an indicator function that counts the number of image-point cloud pairs with an RMSE below the threshold $\tau_6$. The RMSE for each pair $(\mathbf{I}, \mathbf{P}) \in \mathcal{D}$ is calculated as:

\begin{equation}
   \text{RMSE}(\mathbf{I}, \mathbf{P}) = \sqrt{\frac{1}{|\mathbf{P}|}\sum_{\mathbf{x_i} \in \mathbf{P}} \|\Bar{\mathcal{T}}_{\mathbf{I}}^{\mathbf{P}}(\mathbf{x_i}) - \mathcal{T}_{\mathbf{I}}^{\mathbf{P}}(\mathbf{x_i})\|^2},
\end{equation}

where $\mathcal{T}_{\mathbf{I}}^{\mathbf{P}}$ represents the predicted transformation from $\mathbf{I}$ to $\mathbf{P}$, and $\Bar{\mathcal{T}}_{\mathbf{I}}^{\mathbf{P}}$ denotes the ground truth transformation from $\mathbf{I}$ to $\mathbf{P}$. This metric provides an indication of the precision of the cross-modal registration process across the entire dataset.

\section*{B. Detailed Introduction and Discussion on Related Works}
\label{baseline discussion}

\subsection*{B.1. Baseline}
\label{baseline}
\textbf{2D3D-MATR.} 2D3D-MATR \cite{2d3dmatr} is a detection-free method for accurate and robust image-to-point cloud registration. It adopts a coarse-to-fine manner where it first forms a coarse correspondence set between downsampled patches of the input image and the point cloud, then extends them into dense correspondences within the patch. In coarse-level matching, a transformer facilitates contextual sharing between image and point cloud features. Then a multi-scale feature matching module is designed to match each point patch with its most suitable zoomed image patch to avoid the scale ambiguity problem. Finally, the PnP-RANSAC is applied to the fine-level dense correspondences to estimate the transformation. Though the design of the transformer block and the multi-scale matching improve the quality of the extracted correspondences and contribute to accurate 2D-3D registration, the registration process remains hindered by the inherent modality gap between images and point clouds. This gap often results in poor feature matching accuracy, ultimately leading to registration failures.

\subsection*{B.2. Closely Related Work}
\label{other related works}

\textbf{FreeReg.} FreeReg \cite{freereg} adopts a pretrained diffusion model with monocular depth estimators for cross-modality feature extraction. Specifically, it constructs two types of features for establishing correspondences: diffusion features and geometric features. The diffusion features are the intermediate representations of the depth-controlled diffusion model, which shows strong consistency across RGB images and depth maps. The geometric features capture distinct local geometric details on the RGB image and depth map using a monocular depth estimator. The combination of these two features enables accurate cross-modal correspondence estimation for registration. However, it still heavily relies on the explicit feature of the pretrained depth-controlled diffusion model, requiring manual selection of the feature layers. Additionally, its computational cost is significantly higher.
\noindent
\textbf{VP2P-Match.} VP2P-Match primarily focuses on registration in outdoor scenes, with point clouds mainly captured by LiDAR, which differs from the benchmarks used by other baseline methods and our approach. VP2P-Match \cite{differentiable} propose to learn a structured cross-modality latent space to represent pixel features and 3D features via a differentiable probabilistic PnP solver. Specifically, it designs a triplet network to learn VoxelPoint-to-Pixel matching, where the 3D elements are represented using both voxels and points, enabling learning of the cross-modality latent space with pixels. The entire framework is trained end-to-end by applying supervision directly to the predicted pose distribution using a probabilistic PnP solver. Although using VoxelPoint for 3D feature extraction provides more descriptive local features, the registration still fails to bridge the modality gap. Note that VP2P-Match still follows the dense matching convention while leveraging the Monto Carlo strategy to approximate the KL divergence loss of the predicted pose distribution and ground truth pose distribution. It may require more computational overhead to achieve the differentiability, and the large search space of dense matching makes it prone to difficulty in finding the correct correspondences.

\section*{C. Datasets}
\label{datasets}
We train and evaluate \paper\ on two indoor datasets 7-Scenes \cite{7-Scenes} and RGB-D Scenes V2 \cite{rgbdscenesv2}, and compare it with the baselines. We also provide simple evaluation of \paper\ on KITTI \cite{kitti}, which contains dynamic outdoor scenarios. The detailed information are as follows.
\subsection*{C.1 7-Scenes}
The 7-Scenes dataset \cite{7-Scenes} contains RGB-D scans of seven indoor scenes: \textit{Chess, Fire, Heads, Office, Pumpkin, Kitchen, and Stairs}. Each scene includes multiple sequences. We use preprocessed data from \cite{2d3dmatr}, with the preprocessing steps as follows. For each scene, we select 25 consecutive depth maps to generate a dense point cloud, which is then downsampled using a voxel size of 2.5 cm. After generating the point cloud data, we extracted each point cloud's first frame’s corresponding image to form an image-point cloud pair. This process is repeated to generate the entire dataset. After data generation, a selection step is applied: each image is unprojected into 3D space to create a virtual point cloud, and its overlap ratio with the actual point cloud is calculated. Pairs with an overlap ratio below 50\% are removed.
The final dataset includes 2,304 test samples and 5,059 training samples, with the training data further split into 80\% for training and 20\% for validation. Since the images and depth maps in the 7-Scenes dataset are not calibrated, we follow \cite{sanet} by rescaling the images by a factor of $\frac{585}{525}$ to achieve an approximate calibration.

\subsection*{C.2 RGB-D Scenes V2}
The RGB-D Scenes V2 dataset \cite{rgbdscenesv2} contains 14 indoor scenes, labeled from \textit{Scene-1} to \textit{Scene-14}. We follow the same data generation method as for 7-Scenes, but in this dataset, we remove image-point cloud pairs with an overlap ratio below 30\%. Compared to 7-Scenes, RGB-D Scenes V2 has a smaller data volume, so we increase the proportion of training data accordingly. The dataset was randomly split into training, validation, and test sets, containing 1,978, 117, and 386 samples, respectively.

\subsection*{C.3 KITTI}
The KITTI-DC dataset \cite{kitti} contains dynamic image-point cloud pairs. The sparse point clouds are obtained with a 64-line LiDAR scan. The distance between image and point cloud pair is less than 10 meters. We follow RreeReg \cite{freereg} Kitti benchmark for evaluation.

\section*{D. Implementation}
\label{implementation_details}

\subsection*{D.1. Differentiable BPnP Solver}
BPnP \cite{bpnp} efficiently derives accurate gradients of the PnP solver based on the Implicit Function Theorem \cite{bpnp} with excellent numerical stability, we employ it as the differentiable PnP solver in our pipeline. Following BPnP, we first construct the constraint function as $ \mathbf{f}(\mathbf{x}, \mathbf{y}, \mathcal{T}^\prime, \mathbf{K}) = [f_1, f_2, ..., f_m]^T $, where $ \mathbf{x} $ and $ \mathbf{y} $ are the input correspondences, $ \mathcal{T}^\prime = [\mathbf{R}^\prime|\mathbf{t}^\prime]$ is the predicted transformation, $m$ is the number of its variables, and K is the camera intrinsic matrix.
And for all $ i \in \{1, ..., m\}$, $ f_i $ is defined as
\begin{equation}
    f_i = \frac{\partial \sum_{i=1}^n \|\mathcal{K}(\mathbf{R}^\prime \mathbf{x}_i + \mathbf{t}^\prime) - \mathbf{y}_i\|_2^2}{\partial \mathcal{T}^\prime_i} .
\end{equation}
Then given the output gradient $ \nabla\mathbf{z} $, the input gradients $ \nabla\mathbf{x} $ and $ \nabla\mathbf{y} $ can be derived as
\begin{align}
    \nabla\mathbf{x} = \Big[-(\frac{\partial\mathbf{f}}{\partial\mathcal{T}^\prime})^{-1}\frac{\partial\mathbf{f}}{\partial\mathbf{x}}\Big]^{T}\nabla\mathbf{z},
    \\
    \nabla\mathbf{y} = \Big[-(\frac{\partial\mathbf{f}}{\partial\mathcal{T}^\prime})^{-1}\frac{\partial\mathbf{f}}{\partial\mathbf{y}}\Big]^{T}\nabla\mathbf{z}.
\end{align}

\subsection*{D.2. Depth Densification}
The sparse depth projected from the point cloud is densified using simple morphology operations like dilation. Specifically, we first invert the depth values below a threshold (0.1) to match a reference maximum depth value (15.0), enabling a more consistent interpolation. Then a diamond-shaped dilation with a size of 7$ \times $7 is utilized to fill empty areas while preserving significant depth features. After that, Hole closing is performed using erosion and dilation operations with smaller kernels (3$\times$3 and 5$\times$5) to clean up the depth map and remove noise or isolated points. Finally, a median blur and a Gaussian blur filter are applied with a kernel size of 5 to help smooth out the depth map. The depth map is inverted again to return to the original depth scale, ensuring that all valid depth values correspond to real-world distances.

\subsection*{D.3. Loss Functions}
Here we provide the detailed calculation of the circle loss. Following 2D3D-MATR \cite{2d3dmatr}, we define the general circle loss $ \mathcal{L}_i $ of an anchor descriptor $ \mathbf{d}_i $ as:
\begin{equation}
    \mathcal{L}_i=\frac{1}{\delta} \log \left[1+\sum_{\mathbf{d}_j \in \mathcal{D}_i^{\mathcal{P}}} e^{\beta_p^{i, j}\left(d_i^j-\Delta_p\right)} \cdot \sum_{\mathbf{d}_k \in \mathcal{D}_i^{\mathcal{N}}} e^{\beta_n^{i, k}\left(\Delta_n-d_i^k\right)}\right],
\end{equation}
where $ \mathcal{D}_i^{\mathcal{P}} $ and $ \mathcal{D}_i^{\mathcal{N}} $ are the descriptors of its positive and negative pairs, $d_i^j$ is the $\ell_2$ feature distance, $\beta_p^{i, j}=\delta \lambda_p^{i, j}\left(d_i^j-\Delta_p\right)$ and $\beta_n^{i, k}=\delta \lambda_n^{i, k}\left(\Delta_n-d_i^k\right)$ are the individual weights for the positive and negative pairs, where $\lambda_p^{i, j}$ and $\lambda_n^{i, k}$ are the scaling factors for the positive and negative pairs. 

We follow the hyperparameter configuration in \cite{2d3dmatr}. On the coarse level, we generate the ground truth based on bilateral overlap. A patch pair is considered positive if the 2D and 3D overlap ratios between them are both at least 30\%, and negative if both overlap ratios are below 20\%. The overlap ratio between the 2D and 3D patches is used as $\lambda_p$, and $\lambda_n$ is set to 1. On the fine level, a pixel-point pair is positive if the 3D distance is below 3.75 cm and the 2D distance is below 8 pixels, and negative if the 3D distance is above 10 cm or the 2D distance exceeds 12 pixels. The scaling factors are all set to 1. All other pairs are ignored during training on both levels as the safe region. The margins are set to $\Delta_p=0.1$ and $\Delta_n=1.4$.

\begin{figure*}[t]
\centering
\vspace{-12mm}
\includegraphics[width=\textwidth]{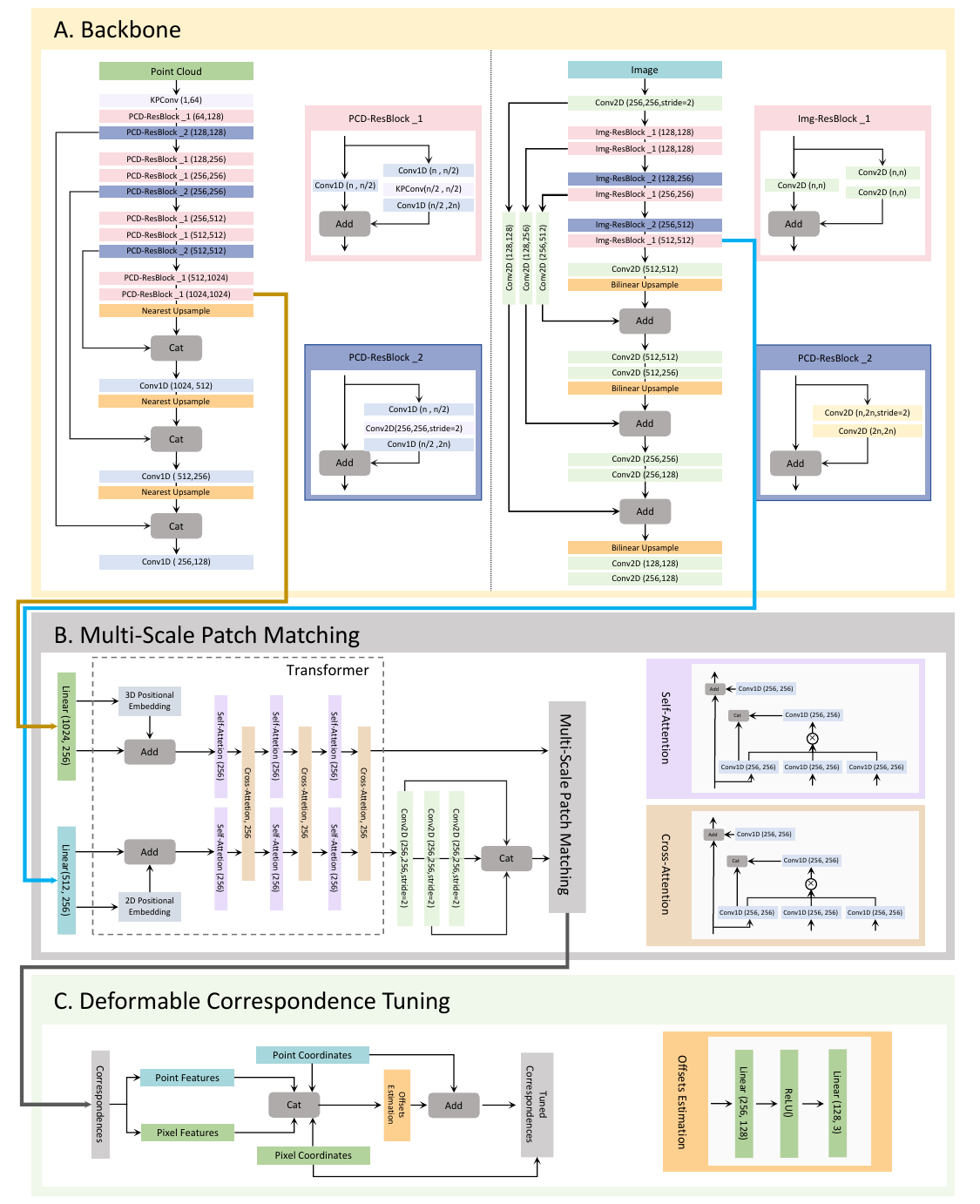}
\captionof{figure}{
The network architecture of our proposed \paper.}
\label{supp_fig:network_architecture}
\end{figure*}

\section*{E. Additional Experiments}
\label{additional_experiments}
\begin{table}[t]
\centering
\caption{The comparison of RRE and RTE results between \paper\ and 2D3D-MATR \cite{2d3dmatr} on the 7-Scenes \cite{7-Scenes} dataset.}
\label{tab:rre_rte}
\begin{tabular}{l|cc}
\toprule
\textbf{Method} & \textbf{RRE (m)} & \textbf{RTE (m)} \\
\midrule
2D3D-MATR \cite{2d3dmatr} & 3.053 & 0.072 \\
\paper\ (ours) & 2.743 & 0.065 \\
\bottomrule
\end{tabular}
\end{table}

\subsection*{E.1. Relative Rotation Error and Relative Translation Error}
Relative Rotation Error (RRE) and Relative Translation Error (RTE) are commonly used to assess the alignment accuracy between two point clouds. In cross-modal registration tasks, these metrics can also evaluate the alignment between a point cloud and an image. Specifically, the input point cloud and the point cloud projected from the depth map corresponding to the image are treated as the two point clouds for evaluation, and their RRE and RTE are computed. \cref{tab:rre_rte} shows the RRE and RTE results of \paper\ compared to the baseline, 2D3D-MATR, where the evaluation dataset is 7-Scenes \cite{7-Scenes}.

As shown in the table, \paper\ significantly outperforms the baseline method in both RRE and RTE metrics. This demonstrates that our method not only accurately estimates the transformation in most scenarios but also achieves high-quality results, ensuring tight alignment between the point cloud and the image.

\subsection*{E.2. Noise and sparse data conditions}
To validate the robustness of \paper\ under noisy scenarios and sparse data conditions, we simulate these conditions by randomizing point coordinates and removing a subset of points. As shown in \cref{tab:noise}, \paper\ maintains stable performance in these simulated scenarios.

\begin{table}[t]
\centering
\caption{Registration under noise and sparse data conditions. Random shifts are sampled from a normal distribution $\mathcal{N}$(0, 0.1) (m).}
\label{tab:noise}
\footnotesize  
\setlength{\extrarowheight}{-7pt}  
\resizebox{\columnwidth}{!}{
\begin{tabular}{l|ccc}
\toprule
\textbf{conditions} & \textbf{IR (\%)$\uparrow$} & \textbf{FMR (\%)$\uparrow$} & \textbf{RR (\%)$\uparrow$} \\
\midrule
(a) random shifts for all points & 53.1 & 91.8 & 82.5 \\
(b) randomize 1\% points' coordinates & 52.7 & 91.2 & 82.0 \\
(c) random remove 10\% points & 53.0 & 91.8 & 81.8 \\
(d) w/o additional conditions & 53.2 & 92.1 & 83.0 \\
\bottomrule
\end{tabular}
}
\end{table}

\subsection*{E.3. Outdoor and dynamic scenarios evaluation}
Our method has been primarily evaluated in static scenarios. To further assess its performance in dynamic and outdoor environments, we conduct experiments on the KITTI dataset. As shown in \cref{tab:kitti}, \paper\ outperforms baselines across nearly all metrics while maintaining an inference speed comparable to that of the fastest method, 2D3D-MATR \cite{2d3dmatr}.

\subsection*{E.4. Comparison with Diff-Reg}
\begin{figure}[h]
    \centering
    \includegraphics[width=\columnwidth]{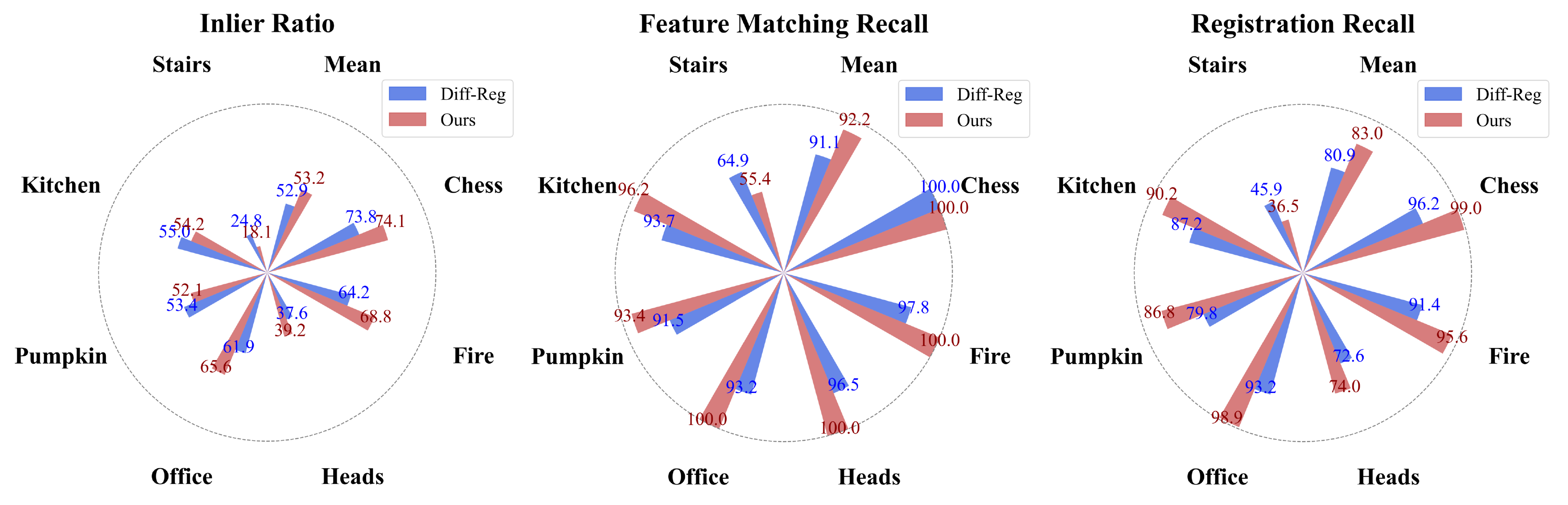}
    \caption{\footnotesize Detailed comparison with Diff-Reg.}
    \label{fig:diffreg}
\end{figure}

Diff-Reg \cite{diffreg} serves as a baseline method that performs the diffusion denoising process at the correspondence set. Although it is not specifically designed for image-to-point cloud cross-modal registration, it can still be applied to this task. Here, we provide a detailed comparison with Diff-Reg. We continue to use 7Scenes as the primary benchmark dataset and report three key metrics: IR (Inlier Ratio), FMR (Feature Matching Recall), and RR (Registration Recall).
We train Diff-Reg on 7-Scenes and select the best-performing checkpoint on the validation set for testing as shown in Fig~\ref{fig:diffreg}.

\subsection*{E.5. Detailed comparison with FreeReg}
\begin{figure}[h]
    \centering
    \includegraphics[width=\columnwidth]{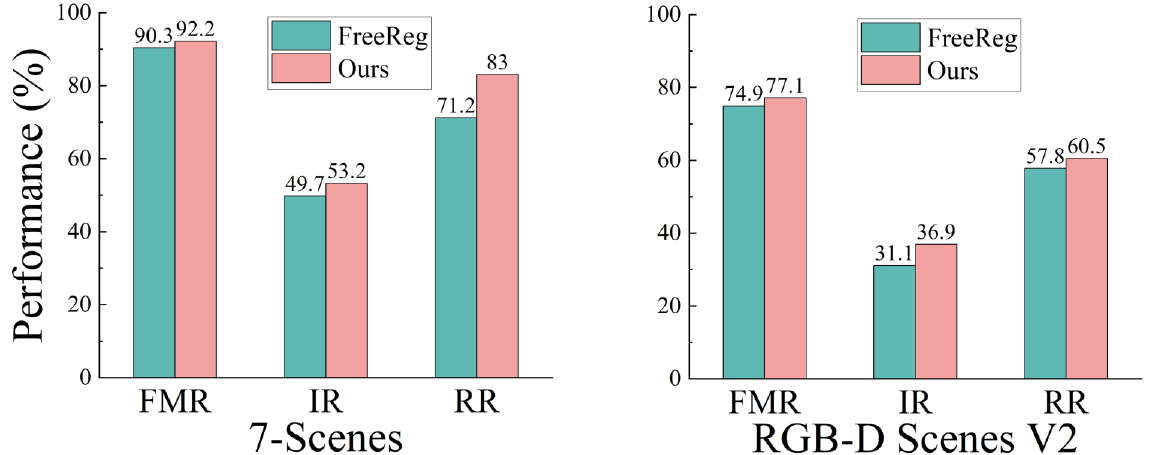}
    \caption{\footnotesize Detailed comparison with FreeReg.}
    \label{fig:freereg}
\end{figure}

Owing to space constraints, we present a simple comparison with FreeReg in the main paper. Additional detailed results on 7-Scenes and RGB-D Scenes V2 are provided in Fig~\ref{fig:freereg}.

\begin{table}[t]
\centering
\caption{Comparison experiments with baselines on KITTI.}
\label{tab:kitti}
\resizebox{\columnwidth}{!}{
\begin{tabular}{l|cccccc}
\toprule
\textbf{Method}& \textbf{FMR (\%)$\uparrow$}& \textbf{IR (\%)$\uparrow$} & \textbf{RRE (°)$\downarrow$}& \textbf{RTE (m)$\downarrow$} & \textbf{RR (\%)$\uparrow$} & \textbf{Time (s)$\downarrow$}\\
\midrule
2D3D-MATR \cite{2d3dmatr} & \textbf{99.7} & 59.1 & 3.334 & 0.838 & 75.4 & \textbf{0.061} \\
FreeReg \cite{freereg} & \textbf{99.7} & 58.3 & 5.987 & 2.414 & 70.5 & 8.763\\
\paper\ (ours) & \textbf{99.7} & \textbf{62.9} & \textbf{2.836} & \textbf{0.773} & \textbf{82.2} & 0.062 \\
\bottomrule
\end{tabular}
}
\end{table}

\section*{F. Visualizations}
\label{visualization}

\begin{figure*}[t]
\centering
\includegraphics[width=\textwidth]{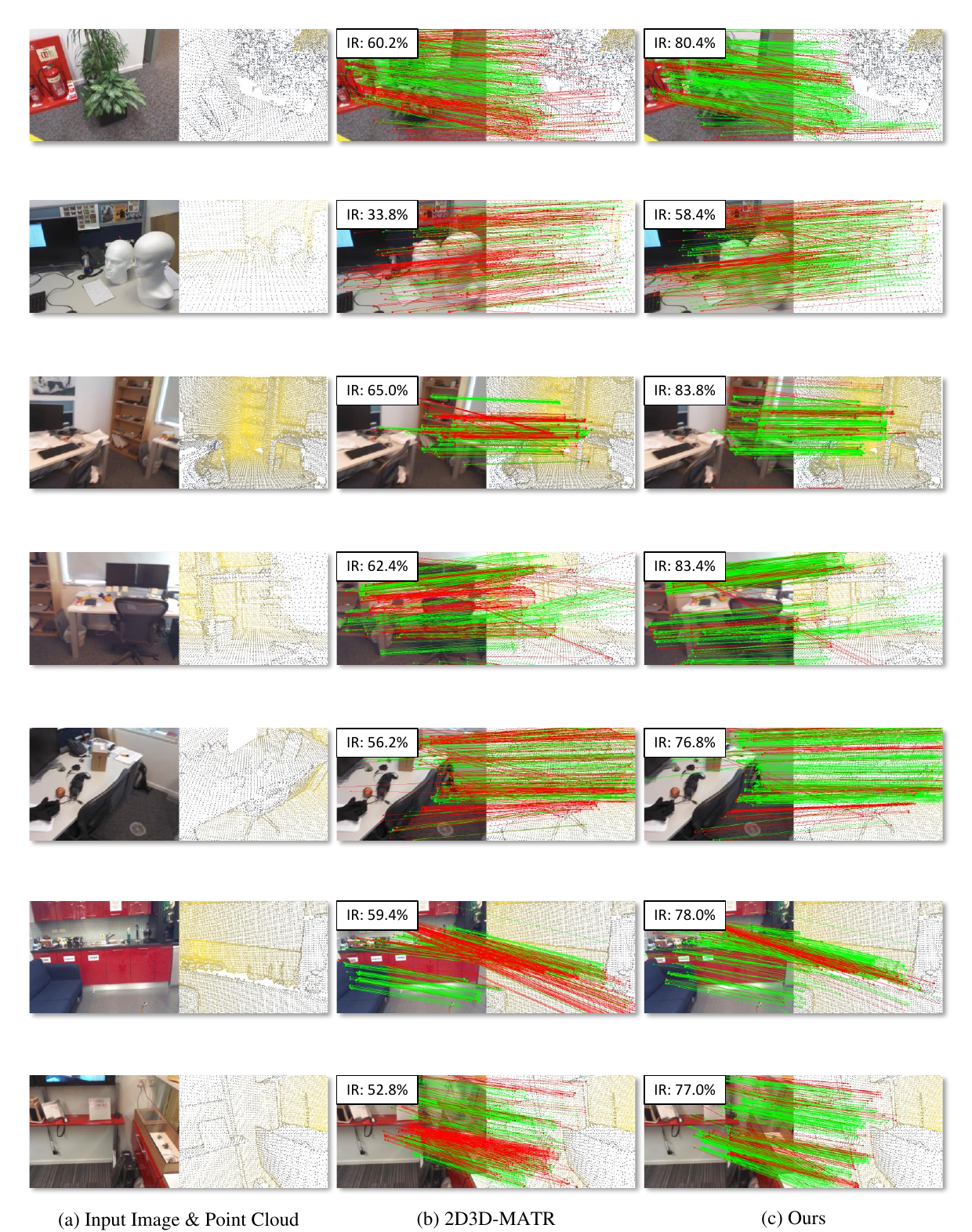}
\captionof{figure}{
Correspondence visualizations on 7-scenes}
\label{supp_fig:qualitative_7scenes}
\end{figure*}

\begin{figure*}[t]
\centering
\includegraphics[width=\textwidth]{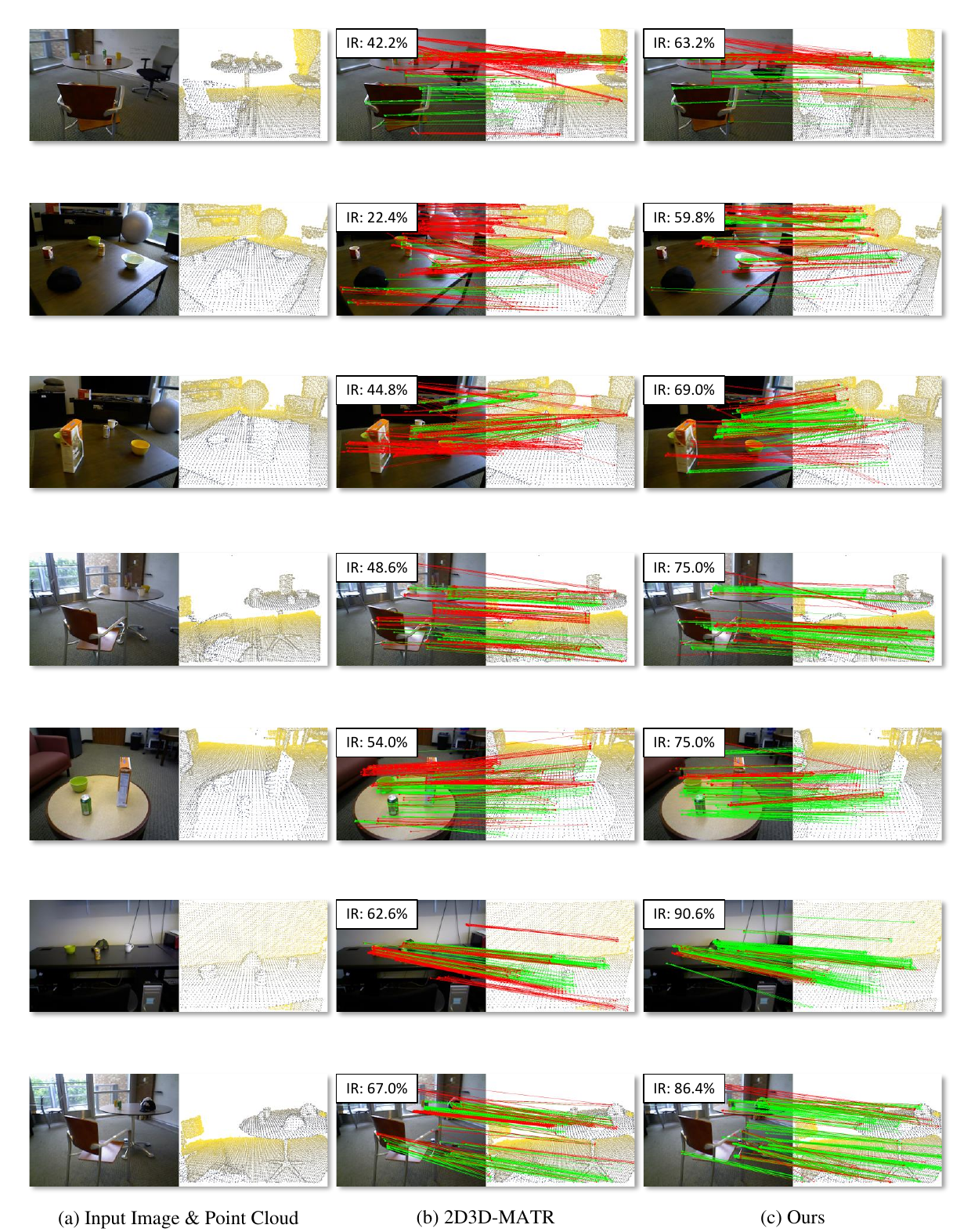}
\captionof{figure}{
Correspondence visualizations on RGB-D Scenes V2.}
\label{supp_fig:qualitative_rgbd}
\end{figure*}

We present additional qualitative results to compare \paper\ with the baseline method, 2D3D-MATR \cite{2d3dmatr}. For clarity, we visualize the correspondences extracted by both methods, selecting the top 500 correspondences with the highest feature matching scores. \cref{supp_fig:qualitative_7scenes} illustrates the results on the 7-Scenes \cite{7-Scenes} dataset, while \cref{supp_fig:qualitative_rgbd} highlights the results on the RGB-D Scenes V2 \cite{rgbdscenesv2} dataset. The findings show that \paper\ extracts more accurate correspondences, delivering robust and superior scene-agnostic registration performance.

\end{document}